\documentclass[10pt,journal,onecolumn] {IEEEtran}

\usepackage{cite}

\usepackage{graphicx}

\usepackage{caption}
\usepackage{subcaption}

\usepackage{amsmath,amssymb,times, rawfonts}
\usepackage{mathrsfs}
\usepackage{dsfont}
\usepackage{array}
\usepackage{authblk}
\usepackage{algorithm,algorithmic}
\usepackage{arydshln}
\newcommand{\deliver}{{\bf deliver}}
\newcommand{\given}{{\bf given}}

\newtheorem{lemma}{Lemma}[section]
\newtheorem{theorem}{Theorem}

\newcommand{\bsb}{\boldsymbol}
\newcommand{\bsbX}{{\boldsymbol{X}}}

\newcommand{\bsby}{{\boldsymbol{y}}}

\newcommand{\bsbb}{{\boldsymbol{\beta}}}
\newcommand{\bsbg}{{\boldsymbol{\gamma}}}

\newcommand{\bsbI}{{\boldsymbol{I}}}

\newcommand{\bsbSig}{{\boldsymbol{\Sigma}}}
\newcommand{\bsbxi}{{\boldsymbol{\xi}}}
\newcommand{\bsbzeta}{{\boldsymbol{\zeta}}}

\newcommand{\bsbV}{{\boldsymbol{V}}}

\newcommand{\bsbe}{{\boldsymbol{e}}}

\newcommand{\bsbu}{{\boldsymbol{u}}}
\newcommand{\bsbv}{{\boldsymbol{v}}}

\newcommand{\rd}{\,\mathrm{d}}
\newcommand{\bsbs}{{\boldsymbol{s}}}
\newcommand{\bsbS}{{\boldsymbol{S}}}

\DeclareMathAlphabet{\mathpzc}{OT1}{pzc}{m}{it}

\begin{document}

\title{
Group Iterative Spectrum Thresholding for Super-Resolution Sparse Spectral Selection}

\author{
\IEEEauthorblockN{Yiyuan She, Jiangping Wang, Huanghuang Li,  and Dapeng Wu}
}


\maketitle
\thispagestyle{empty}

\begin{abstract}
Recently, sparsity-based algorithms are proposed for super-resolution spectrum estimation. However, to achieve adequately high resolution in real-world signal analysis, the dictionary atoms have to be close to each other in frequency, thereby resulting in a coherent design. The popular convex compressed sensing methods break down in  presence of  high  coherence  and large noise. We propose a new regularization approach to handle  model collinearity and  obtain parsimonious frequency selection simultaneously. It takes advantage of the pairing structure of sine and cosine atoms in the frequency dictionary. A probabilistic spectrum screening is also developed for fast computation in high dimensions. A data-resampling version of high-dimensional Bayesian Information Criterion is used to determine the regularization parameters. Experiments show the efficacy and efficiency of the proposed algorithms in  challenging situations with small sample size, high frequency resolution, and low signal-to-noise ratio.

\end{abstract}


\section{Introduction}
\label{sec:Introduction}
The problem of spectral estimation studies how  signal power is distributed over frequencies, and has rich applications in speech coding, radar \&  sonar signal processing and many other areas. 
Suppose a discrete-time real-valued signal is observed at finite time points  contaminated with i.i.d. Gaussian noise.
In common with all spectral models, we assume the signal can be represented as a linear combination of sinusoids, and aim to recover the spectrum of the signal at a desired   resolution.  However, the problem becomes very challenging when the required frequency resolution is high. In particular, the number of the frequency levels at the desired resolution can be (much) greater than the sample size,  referred to as {super-resolution} spectral estimation.
For such discrete-time signals of finite length, the classical methods based on fourier analysis \cite{stoica2005spectral} or least-squares periodogram (LSP) \cite{stoica2009spectral,lomb1976least} suffer from power leakage and have very limited  spectral resolution \cite{stoica2005spectral}.
Some more recent algorithms, such as  Burg \cite{stoica2005spectral},  MUSIC \cite{schmidt1986multiple} and RELAX  \cite{li2002efficient} only alleviate the issue to some extent. 

We assume that the signal is sparse in the frequency-domain, i.e., the number of its sinusoidal components is small relative to the sample size, referred to as the \textit{spectral sparsity}. It is a realistic assumption in many applications (e.g.,  astronomy \cite{chen1998application}  and radar signal processing \cite{li2009mimo}), and makes it possible to apply the revolutionary \textbf{compressed sensing} (CS) technique. In \cite{chen1998application}, Chen and Donoho  proposed the basis pursuit (BP) to handle overcomplete dictionaries and unevenly sampled signals. A number of similar works followed, see, e.g.,  \cite{bourguignon2007sparsity,candes2008enhancing,fuchs2002minimal,sparspec,blumensath2009iterative,blumensath2009normalised,lza-f,isl0} and the references therein.

We point out   two crucial facts that cannot be ignored in super-resolution spectrum reconstruction. (a) When the desired frequency resolution is very high, neighboring dictionary atoms become very similar and thus necessarily result in high coherence  or collinearity. As is well known in the  literature, the popular convex $l_1$ technique 
as used in the BP  yields inconsistent frequency selection and suboptimal rates in estimation and prediction under such coherent setups \cite{donoho2006stable,candes2006stable,zhang2008sparsity,zhao2006model,candplan09}. 
(b) The grouping structure of the sinusoidal components is an essential feature in spectrum recovery: if frequency $f$  is absent in the signal, the coefficients for $\cos(2\pi f t)$ and $\sin(2\pi f t)$ should \emph{both} be zero.

In this paper we investigate super-resolution spectral recovery from a statistical perspective and propose a \textbf{group iterative spectrum thresholding} (\textbf{GIST}) framework to tackle the aforementioned challenges. GIST  allows for (possibly nonconvex) shrinkage estimation and can exploit the pairing structure. 
Interestingly, we find that neither the $l_1$ nor the $l_0$ regularization is satisfactory for spectrum estimation, and advocate a hybrid   $l_0+l_2$ type shrinkage estimation.  Theoretical analysis shows that the new regularization  essentially removes the stringent coherence requirement and  can accommodate much lower SNR and higher coherence.
Furthermore, a GIST variant provides a screening technique for supervised dimension reduction to deal with applications in ultrahigh dimensions. The rest of this paper is organized as follows. We formulate the  problem from a statistical point of view and briefly survey the literature in Section \ref{sec:Survey}. In Section \ref{sec:Model},  we  propose  the GIST framework---in more details,  a novel form of regularization,  a generic  algorithm for fitting  group nonconvex penalized models, a data-resampling based model   selection criterion, and  a probabilistic spectral screening for fast computation.
Experimental results are shown in Section \ref{sec:Experiments}. We summarize the conclusions  in Section \ref{sec:Conclusions}.
The technical details are left to the Appendices.

\section{Model Setup and The Super-resolution Challenge}
\label{sec:Survey}

In this section, we introduce the problem of super-resolution spectrum estimation and  review some existing methods from a statistical point of view.
Let  $\bsby=[y(t_n)]_{1\leq n\leq N}$ be a \textbf{real}-valued signal contaminated with i.i.d. Gaussian noise $N(0, \sigma^2)$. (We focus on real signals in this paper but our methodology carries over to complex-valued signals; see Section \ref{sec:Conclusions}.)
The sampling time sequence $\{t_n\}_{1\leq n \leq N}$ is \emph{not} required to be uniform  (cf. \cite{chen1998application}). In order to achieve super-resolution spectral recovery, an \emph{overcomplete} frequency dictionary must be applied.
Concretely, we use a grid of evenly spaced frequencies $f_k = f_{\max} \cdot k/D$ for $k=0, 1, \cdots, D$ to construct the sine and cosine frequency predictors, i.e., $\cos(2\pi t  f_k)$ 
and $\sin(2\pi t  f_k)$. Let $\mathcal F$ denote the set of  nonzero frequencies $\{f_1, \cdots, f_D\}$.
The upper band limit $f_{\max}$ can be $(2 \min_{1\leq n \leq N}(t_n - t_{n-1}))^{-1}$  or estimated based on the spectral window~\cite{scargle1982studies}.
The cardinality of the dictionary controls the frequency resolution  given by $f_{\max}/D$.
The true spectra of the signal are assumed to be discrete for convenience, because the   quantization error can always be reduced by increasing the value of $D$. The signal can be  represented by
\begin{eqnarray}
y_n=y(t_n) = \sum_{k=0}^{D}A_k\cos(2\pi f_k t_n+\phi_k) + e_n, 1\leq  n \leq N, \label{origsignalmodel}
\end{eqnarray}
where $A_k$, $\phi_k$ are unknown, and the noise $\{e_n\}_{n=1}^N$ are i.i.d.   Gaussian with zero mean and unknown variance $\sigma^2$.
Traditionally, $D\leq N$. But in super resolution spectral analysis,  $D$ can take a much larger value than $N$. It still results in  a well-defined problem because only a few $A_k$ are nonzero under the \textit{spectral sparsity} assumption.

From $A_k\cos(2\pi f_kt_n+\phi_k)$
$= A_k\cos(\phi_k)\cos(2\pi f_kt_n)-A_k\sin(\phi_k)\sin(2\pi f_kt_n)$
$= a_k\cos(2\pi f_kt_n)+b_k\sin(2\pi f_kt_n)$ with $a_k = A_k\cos \phi_k$, $b_k = -A_k\sin \phi_k$, we introduce two column vectors
\begin{align*}
\bsbX^{\cos} (f) \triangleq \left[\cos(2\pi t_n  f) \right]_{1\leq n \leq N}, \\
  \bsbX^{\sin}(f)\triangleq\left[\sin(2\pi t_n  f)\right]_{1\leq n \leq N},
\end{align*}
and define the predictor matrix
\begin{align}
\bsbX \triangleq [\bsbX^{\cos}(f_1), \cdots, \bsbX^{\cos}(f_D),    \bsbX^{\sin}(f_1), \cdots, \bsbX^{\sin}(f_D) ]. \label{eqn:X0}
\end{align}
 (Some redundant or useless predictors can be removed in concrete problems, see \eqref{eqn:X}.)
Denote the coefficient vector by $\bsbb\in {\mathbb R}^{2D}$  and the intercept (zero frequency component) by $\alpha$.
Now the model can be formulated as a linear regression
\begin{align}
\bsby  = \alpha + \bsbX\bsbb + \bsb{e}, \label{ols}
\end{align}
where $\bsbb$ is sparse and $\bsb{e} \sim N(\mathbf{0},\sigma^2\bsbI)$. 
In super-resolution analysis,  $D\gg N$, giving a small-sample-size-high-dimensional design.
Linear analysis such as Fourier transform fails for such an {underdetermined} system.

As a demonstration, we consider a noisy `TwinSine' signal at frequencies $0.25$ Hz and $0.252$ Hz with 100 observations. 
Obviously, the frequency resolution needs to be as fine as $0.002$ HZ to perceive and distinguish the two sinusoidal components with different coefficients. We set $f_{\max}=1/2$,
and thus $2D$ must be at least $500$ -- much larger than the sample size.
The concrete design matrix (without the intercept) is given by
\begin{equation}\label{eqn:X}
\bsbX =\left[
\begin{smallmatrix}
\cos( \pi\frac{1}{D}t_1) &  \cdots & \cos( \pi\frac{D}{D}t_1) & \sin( \pi\frac{1}{D}t_1) &  \cdots & \sin( \pi\frac{D-1}{D}t_1) \\
 \vdots & \vdots & \vdots\quad & \vdots & \vdots & \vdots \\
\cos( \pi\frac{1}{D}t_N) &  \cdots & \cos( \pi\frac{D}{D}t_N) & \sin( \pi\frac{1}{D}t_N) &  \cdots & \sin( \pi\frac{D-1}{D}t_N) \\
\end{smallmatrix}
\right].
\end{equation}
The last  sine atom  disappears because all $t_n$ are integers.
This yields a super-resolution spectral estimation problem. 

There are many algorithms for identifying the spectrum of a discrete-time signal. But not all of them can super-resolve. From a modeling perspective, we classify them as nonsparse methods and sparse methods.
Most classical methods (e.g., \cite{lomb1976least,scargle1982studies,stoica2009spectral})  are nonsparse and assume  no knowledge on the power spectra. For super-resolution spectrum estimation, they  may seriously broaden the main lobes and introduce side lobes. \emph{In this paper, we focus on sparse methods.}
 As aforementioned, one popular assumption for solving underdetermined systems is signal sparsity:  
the number of present frequency components  is small relative to the number of samples.
The problem is still NP hard because the frequency location of  the truly relevant sinusoidal components is unknown and the number of candidate components can be very large. In fact, the frequency grid used for constructing the dictionary can be made arbitrarily fine by the customer.

Early attempts to enforce sparsity effects include greedy or  exhaustive searches \cite{mallat1993matching,natarajan1995sparse} and genetic algorithms with a sparsity constraint \cite{holland1992genetic}. Harikumar \cite{harikumar1996new} computes the maximally sparse solutions under a constraint on the fitting error. 
A breakthrough is due to Chen \& Donoho  who proposed  the basis pursuit (BP) for spectrum estimation \cite{chen1998application}. A number of similar works followed \cite{sparspec,bourguignon2007sparsity,candes2008enhancing,fuchs2002minimal}. BP is able to superresolve  for unevenly sampled signals.
In our notation, the  noiseless version of BP solves the convex optimization problem
$    \min ||\bsbb||_1 \text{  s.t.  } \alpha + \bsbX \bsbb = \bsby.
$
The noisy versions can be defined similarly, in a penalty/constraint form.
The $l_1$-norm provides the tightest convex relaxation to the $l_0$-norm and achieves a sparse spectral representation of the signal within feasible time and cost.

In recent years, the power and limitation of this convex relaxation have been systematically studied in a large body of compressed sensing literature. In short, to guarantee good statistical performance in either prediction, estimation, or model selection, the coherence of the system must be low, in terms of, e.g.,  \textit{mutual coherence} conditions \cite{donoho2006stable}, \textit{restricted isometry property} (RIP) \cite{candes2006stable} and \textit{irrepresentable conditions} \cite{zhao2006model} among others. For example, the RIP of order $s$ requires that for any index set $I\subset \mathcal F$ with $|I|=s$,   there exists an RIP constant $\delta_s\geq 0$ such that
$(1-\delta_s) \|\bsbv\|_2^2 \leq \|\bsbX_I \bsbv\|_2^2 \leq (1+\delta_s) \|\bsbv\|_2^2$, $\forall v \in \mathbb R^s$; when $\delta_s$ is small, any $s$ predictors in $\bsbX$ are approximately  orthogonal. In theory,  to guarantee  $l_1$'s effectiveness in statistical accuracy, frequency selection consistency, and algorithmic stability, such RIP constants have to be small, e.g., $\delta_{3S} + 3\delta_{4S} <2$ in a noisy setup, where  $S=\| \bsbb\|_0$  \cite{candes2006stable}. Similarly, the  mutual coherence, defined as the maximum absolute value of the off-diagonal elements in the scaled Gram matrix  $\bsbX^T \bsbX/N$, has to be as low as $O(1/S)$ \cite{donoho2006stable}.
Such theoretical results  clearly indicate that the super-resolution challenge \emph{cannot} be   \textbf{fully} addressed by the $l_1$-norm based methods, because     many similar sinusoidal components may arise in the dictionary and  bring in high coherence.


To enhance the sparsity of the BP,  Blumensath \& Davies
 proposed the iterative hard thresholding (IHT) \cite{blumensath2009iterative,blumensath2009normalised}. 
See    \cite{lza-f,isl0} for some approximation methods.
Intuitively,   nonconvex penalties can better approximate the $l_0$-norm  and yield sparser estimates than the convex $l_1$-penalty. On the other hand, we find that when the signal-to-noise ratio (SNR) is low and/or the coherence is high, the $l_0$ penalization may give an \textbf{over-sparse} spectral estimate and miss certain true frequency components.
The high miss rates are due to the fact that the $l_0$ regularization is through (hard) thresholding only,  offering   {no} shrinkage at all for nonzero coefficients.
Therefore, it tends to kill too many predictors to achieve the appropriate extent of shrinkage especially when the SNR is low.  An inappropriate nonconvex penalty may seriously mask true signal components.
This issue will be examined in the next section.


\section{GIST Framework} 
\label{sec:Model}
This section examines the super-resolution spectrum estimation in details.
The complete group iterative spectrum thresholding (GIST) framework is introduced at the end.

\subsection{A novel regularization form}
\label{subsec:Model}
In this subsection, we study a  group  penalized least-squares model and investigate the appropriate type of regularization.

The BP finds a solution to an underdetermined linear system  with the minimum $l_1$ norm. 
When the signal is corrupted by noise as in \eqref{ols}, the following $l_1$-penalized linear model is more commonly used:
\begin{equation}\label{eqn:L1 Penalized Linear Model}
 {1\over2}\|\bsby - \alpha-\bsbX \bsbb \|_2^2 + \lambda\|\bsbb\|_1,
\end{equation}
where $\lambda$ is a regularization parameter to provide  a trade-off between the fitting error and   solution sparsity. 
The intercept or zero frequency component $\alpha$ is not subject to any penalty.
To include  more sparsity-enforcing penalties, we consider a more general problem in this  paper which minimizes
\begin{equation}\label{eqn:Penalized Linear Model}
 {1\over2}\|\bsby -\alpha- \bsbX \bsbb \|_2^2 + \sum_{k=1}^{2D} P(|\beta_k|; \lambda)=:F(\bsbb;\lambda),
\end{equation}
where $P(\cdot;\lambda)$ is a univariate penalty function parameterized by  $\lambda$ and is possibly \textbf{nonconvex}.

Some structural information can be further incorporated in spectrum estimation.
From the derivation of \eqref{ols}, $A_k=0$ implies $\beta_k=\beta_{D+k}=0$, i.e., the sine and  cosine predictors at $f_k$ vanish \emph{simultaneously}.
The pairing structure shows it is more reasonable to impose the so-called group sparsity or block sparsity \cite{Yuan,Block2009} on $\{(\beta_k, \beta_{D+k})\}_{1\leq k \leq D}$ rather than the unstructured sparsity on $\{\beta_k\}_{1\leq k \leq 2D}$.  The  group penalized model with the model design  \eqref{eqn:X0} minimizes
\begin{equation}\label{eqn:Grouped Penalized Linear Model}
 {1\over2}\|\bsby - \alpha-\bsbX \bsbb \|_2^2 + \sum_{k=1}^D P\left(\sqrt{\beta_k^2+\beta_{D+k}^2}; \lambda\right)=:F(\bsbb;\lambda).
\end{equation}
(In the problem with the design matrix given by \eqref{eqn:X}, the last sine predictor disappears and thus we always set $\beta_{2D}$   to be $0$.)
The penalty function $P$ is the same as before and is allowed to  be nonconvex.
For ease in computation, the first term in \eqref{eqn:Penalized Linear Model} and \eqref{eqn:Grouped Penalized Linear Model} will be replaced by ${1\over2}\|\bsby - \alpha-\bsbX \bsbb \|_2^2/C$ for some $C$ large enough; see the comment after Theorem \ref{th:conv}.

A crucial problem is then to determine the appropriate form of $P$ for regularization purposes.
The popular $l_1$-penalty $P_1(t;\lambda) = \lambda |t|$ may result in insufficient sparsity and relatively large prediction error, as shown in Section \ref{sec:Experiments}. 
There is still much room for improvement in  super-resolution spectral estimation.
Before we proceed, it is worth pointing out that  there are two objectives involved in this task 
\\
\textbf{Objective 1} (\textbf{O1}): \emph{accurate} prediction of the signal at any new time point in the time domain;
\\
\textbf{Objective 2} (\textbf{O2}): \emph{parsimonious}  spectral representation of   the signal in the Fourier domain.
\\
{\bf O1}+{\bf O2} complies with \textbf{Occam's razor} principle---the simplest way to explain the data is the best. A perfect approach must reflect both concerns to produce a stable sparse model with good generalizability.

\begin{figure}[h!]
  \centering
  \includegraphics[width=3.5in, height=2in]{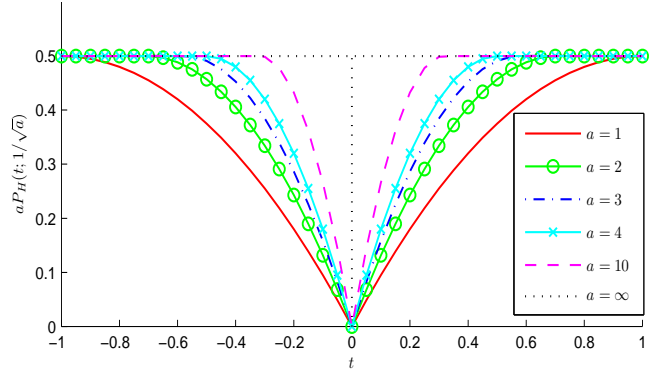}
  \caption[]{{The nonconvex `hard' penalty family ($a\geq 1$) in a neighborhood around $0$. All  penalties lead to the same $\Theta$-estimators. The discrete $l_0$-penalty $P_0$ corresponds to $a =\infty$. The one with the smallest curvature is given by $P_H$ with $a=1$. 
  }    }
  \label{fig:penalties}
\end{figure}

From the perspective of \textbf{O2}, the $l_0$-norm constructs an ideal penalty
\begin{align}
P_0(t;\lambda) = \frac{\lambda^2}{2} 1_{t\neq 0}, \label{l0-pen}
\end{align}
where the indicator function $1_{t\neq 0}$ is $1$ when $t\neq 0$ and $0$ otherwise. Yet it is discrete and strongly nonconvex.
Interestingly, given any model matrix, the class of penalties $a P_H(t;\lambda/\sqrt a)$ for any $a\geq 1$  mimics the behavior of \eqref{l0-pen}, where  $P_H$, referred to as the  \emph{hard-penalty}, is defined by
\begin{eqnarray}
P_H(t; \lambda)= \begin{cases} -t^2/2+\lambda |t|,  & \mbox{ if } |t|<\lambda\\
\lambda^2/2,  & \mbox{ if } |t|\geq \lambda. \end{cases} \label{hard-pen}
\end{eqnarray}
Based on \cite{SheGLMTISP}, we can show that all penalties,  including the continuous penalty \eqref{hard-pen} ($a=1$) and the discrete penalty \eqref{l0-pen} ($a=\infty$), result in the same global minima in optimization.
Fig. \ref{fig:penalties} illustrates the  penalty family in a neighborhood around 0.

A different type of regularization is desirable for objective \textbf{O1}. Even if all truly relevant sinusoidal components could be  successfully located, these atoms are not necessarily far apart in the frequency domain, and thus {collinearity} may occur. 
In statistical signal processing, Tikhonov regularization is an effective means to deal with the singularity issue which seriously affects estimation and prediction accuracy. It is in the form of an $l_2$-norm penalty   
\begin{align}
P_R(t;\eta) = \frac{1}{2} \eta t^2,  \label{l2-pen}
\end{align}
also known as the ridge penalty in statistics. The necessity and  benefit  of introducing such shrinkage in multidimensional estimation date back to the  famous James-Stein estimator  \cite{JamesStein1961}.
Even for the purpose of detection, \textbf{O1} plays an important role because most parameter tuning methods are designed to reduce prediction error.

Taking into account both concerns, we advocate the following hybrid \emph{hard-ridge} (\textbf{HR}) penalty as a fusion of \eqref{hard-pen} and \eqref{l2-pen}:
\begin{align} \label{hrpen}
P_{HR}(t; \lambda,\eta) =
\begin{cases} -\frac{1}{2} t^2 + \lambda |t|, &\mbox{ if } |t| < \frac{\lambda}{1+\eta}\\ \frac{1}{2} \eta t^2 +\frac{1}{2}\frac{\lambda^2}{1+\eta}, &\mbox{ if } |t| \geq \frac{\lambda}{1+\eta}. \end{cases}
\end{align}
The hard portion induces sparsity for small coefficients, while the ridge portion, representing Tikhonov  regularization, helps address the coherence of the design and compensates for noise and collinearity. In the following  subsections, we will show that such defined hard-ridge penalty also allows for ease in optimization and has better frequency selection performance.

Finally, we point out the difference between HR and the elastic net  \cite{zou2005regularization}  which adds an additional ridge penalty in the lasso problem  \eqref{eqn:L1 Penalized Linear Model}. However, this $l_1+l_2$  penalty, i.e., $\lambda_1 \|\bsbb\|_1+\lambda_2^2 \|\bsbb\|_2^2/2,$ may over-shrink the model (referred to as the \emph{double-shrinkage} effect \cite{zou2005regularization}) and can not enforce higher level of sparsity than the $l_1$-penalty.   
In contrast,  using a $q$-function trick  \cite{SheIPOD},  it is shown  that   $P_{HR}$ results in the same estimator as  the `$l_0+l_2$' penalty
\begin{align}
P(t;\lambda, \eta) = \frac{1}{2}\frac{\lambda^2}{1+\eta} 1_{t\neq 0} + \frac{1}{2} \eta t^2.   \label{l02-pen}
\end{align}
The ridge part does not affect the nondifferential behavior  of the $l_0$-norm  at zero, and there is no double-shrinkage effect for nonzero coefficient estimates.


\subsection{GIST fitting algorithm} 
\label{subsec:Solving}

We discuss how to fit the group penalized model  \eqref{eqn:Grouped Penalized Linear Model} for a wide class of penalty functions. We assume both $\bsbX$ and $\bsby$ have been centered so that the intercept term vanishes in the model.
Our  main tool to tackle the computational challenge is the class of $\Theta$-estimators \cite{she2009thresholding}.
Let $\Theta(\cdot;\lambda)$ be an arbitrarily given threshold function (with $\lambda$ as the parameter) which is odd, monotone, and a unbounded shrinkage rule (see \cite{she2009thresholding} for the rigorous definition) with $\lambda$ as the parameter. A group $\Theta$-estimator is defined to be a solution to 
\begin{align}
\bsbb = \vec \Theta ( \bsbb + \bsbX^T (\bsby - \bsbX \bsbb); \lambda). \label{thetaeq-grp}
\end{align}
Here, for any $\bsbxi\in \mathbb R^{2D}$, $\vec \Theta(\bsbxi; \lambda)$  is a $2D$-dimensional vector  $\bsbxi'$ satisfying
$$
[\xi_k ',  \xi_{k+D} ' ]=
[ \xi_k,   \xi_{k+D} ] \Theta\left( \|  [ \xi_k,  \xi_{k+D}  ] \|_2;\lambda\right) /  \|  [ \xi_k,  \xi_{k+D} ] \|_2,
$$
for  $1\leq k \leq D$.
In the simpler case when no grouping is assumed, the $\Theta$-estimator equation \eqref{thetaeq-grp} reduces to
\begin{align}
\bsbb =  \Theta ( \bsbb + \bsbX^T (\bsby - \bsbX \bsbb); \lambda). \label{thetaeq}
\end{align}
A $\Theta$-estimator is necessarily a $P$-penalized estimator provided that
\small
\begin{align}
P(t;\lambda)-P(0;\lambda)=
\int_0^{|t|} (\sup\{s:\Theta(s;\lambda)\leq u\} - u) \rd u  + q(t;\lambda) \label{pen-theta}
\end{align}
\normalsize
holds for some nonnegative $q(\cdot;\lambda)$ satisfying $q(\Theta(s;\lambda);\lambda)=0$ for any $s\in \mathbb R$ \cite{SheGLMTISP}.
Based on this result, we can compute   $P$-penalized estimators by solving \eqref{thetaeq-grp} for an appropriate $\Theta$.


\begin{algorithm}
\begin{algorithmic}
\caption{GIST-fitting algorithm. \label{alg:TISP for PLM}}
\STATE \given\ $\bsbX$ (design matrix, normalized), $\bsby$ (centered),
 $\lambda$ (regularization parameter(s)), $\Theta$ (thresholding rule),
  $\omega$ (relaxation parameter), and $\Omega$ (maximum number of iterations).
\STATE  1) $\bsbX\leftarrow \bsbX/\tau_0$, $\bsby\leftarrow \bsby/\tau_0$, 
with  $\tau_0\geq \|\bsbX\|_2$ (spectral norm).
\STATE 2) Let  $j\leftarrow 0$ and $\bsbb^{(0)}$ be an initial estimate, say, $\bsb{0}$.
\WHILE{$\|\bsbb^{(j+1)}-\bsbb^{(j)}\|$ is not small enough or $j \leq \Omega$}
\STATE
  3.1) $\bsbxi^{(j+1)} \leftarrow (1-\omega)\bsbxi^{(j)}+\omega ( \bsbb^{(j)}+ \bsbX^T(\bsby-\bsbX\bsbb^{(j)}) )$ if $j>0$, and
 $\bsbxi^{(j+1)} \leftarrow    \bsbb^{(j)}+ \bsbX^T(\bsby-\bsbX\bsbb^{(j)})$ if $j=0$;
\STATE
 \underline{{\sc Group form}:}\\
 3.2a)  $l_k^{(j+1)}\leftarrow \sqrt{ (\xi_{k}^{(j+1)} ) ^2 + (\xi_{k+D}^{(j+1)}) ^2}$, $1\leq k \leq D$.\\
 3.2b)  If  $l_k^{(j+1)}\neq 0$, $\bsbb_{k}^{(j+1)}\leftarrow {\xi_{k}^{(j+1)} } \Theta ( l_k^{(j+1)};{\lambda} )/ l_k^{(j+1)}$  and  $\bsbb_{k+D}^{(j+1)}\leftarrow {\xi_{k+D}^{(j+1)}  }\Theta ( l_k^{(j+1)};{\lambda} )/l_k^{(j+1)}$.
 Otherwise   $\bsbb_{k}^{(j+1)}=\bsbb_{k+D}^{(j+1)}=0$.\\
  \underline{{\sc Non-Group form}:}\\
 3.2')     $\bsbb^{(j+1)}\leftarrow \Theta  ( \bsbxi^{(j+1)}; {\lambda}  )$; 
\ENDWHILE
\STATE\deliver\  $\hat{\bsbb}=\bsbb^{(j+1)}$.
\end{algorithmic}
\end{algorithm}

We  establish the convergence of Algorithm \ref{alg:TISP for PLM} in the following theorem. For simplicity, assume that there is no intercept term in the model (which is reasonable when $\bsbX$ and $\bsby$ have both been centered),  and $\tau_0 = 1 > \|\bsbX\|_2$. Let $\bsbSig=\bsbX^T \bsbX$.  Construct an energy function for any $\bsbg, \bsbzeta, \bsbb, \bsbxi\in \mathbb R^{2D}$ as follows

\begin{align}
&G(\bsbg, \bsbzeta, \bsbb, \bsbxi)  =   \frac{1}{2} \| \bsbX \bsbg - \bsby\|_2^2 + P(\bsbg;\lambda) + \frac{\omega}{2} (\bsbg - \bsbb)^T (\bsbI - \bsbSig)(\bsbg-\bsbb)\notag\\
 & +\frac{(1-\omega)^2}{2\omega} (\bsbzeta - \bsbxi)^T (\bsbI - \bsbSig)^{-1} (\bsbzeta - \bsbxi)  +\frac{1-\omega}{2} [\bsbg+ (\bsbI - \bsbSig)^{-1} \bsbX^T \bsby \notag \\
 & -(\bsbI-\bsbSig)^{-1} \bsbxi]^T (\bsbI - \bsbSig) [\bsbg + (\bsbI - \bsbSig)^{-1} \bsbX^T \bsby - (\bsbI-\bsbSig)^{-1} \bsbxi] \notag\\
&+ \frac{1-\omega}{2} [\bsbzeta - (\bsbI-\bsbSig) \bsbb - \bsbX^T \bsby]^T (\bsbI - \bsbSig)^{-1}  [\bsbzeta - (\bsbI-\bsbSig) \bsbb - \bsbX^T \bsby] \notag\\
 &- \frac{1-\omega}{2} [ \bsbxi - (\bsbI - \bsbSig) \bsbb - \bsbX^T \bsby]^T (\bsbI - \bsbSig)^{-1} [\bsbxi - (\bsbI - \bsbSig) \bsbb - \bsbX^T \bsby],\label{defG}
\end{align}
\normalsize
with the non-group and group versions of $P(\bsbg;\lambda)$ being $\sum_{k} P(|\gamma_k|; \lambda)$ and $\sum_{k=1}^D P\left(\sqrt{\gamma_k^2+\gamma_{D+k}^2}; \lambda\right)$, respectively. $G(\bsbg, \bsbzeta, \bsbb, \bsbxi)$ is always greater than or equal to the objective function  $F(\bsbg)$ as defined in \eqref{eqn:Grouped Penalized Linear Model} or \eqref{eqn:Penalized Linear Model}.  This energy function can be used to prove the convergence of the iterates to a $\Theta$-estimator.

\begin{theorem}\label{th:conv}
For any $0< \omega \leq 1$  and a thresholding rule $\Theta$ satisfying \eqref{pen-theta}, under the continuity assumption $\mathfrak{A}$ in Appendix \ref{app:proofconv},  Algorithm \ref{alg:TISP for PLM}  in either  group form or non-group form  converges,  and the iterates $(\bsbb^{(j)}, \bsbxi^{(j)})$ satisfy
\begin{eqnarray}
G(\bsbb^{(j+1)}, \bsbxi^{(j+1)}, \bsbb^{(j+1)}, \bsbxi^{(j+1)}; \lambda) \leq  G(\bsbb^{(j)}, \bsbxi^{(j)}, \bsbb^{(j)}, \bsbxi^{(j)}; \lambda) - \delta_1 - \delta_2, \label{fundecrease}
\end{eqnarray}
 where
$\delta_1 =   \frac{1-\omega}{2\omega} (\bsbxi^{(j+1)} - \bsbxi^{(j)})^T ( \bsbI - \bsbSig)^{-1} (\bsbxi^{(j+1)} - \bsbxi^{(j)})$ and
$\delta_2 =  \frac{1}{2\omega} [\omega ( \bsbI - \bsbSig)  (\bsbb^{(j)} - \bsbb^{(j+1)}) + (1-\omega) (\bsbxi^{(j)} - \bsbxi^{(j+1)})]^T ( \bsbI - \bsbSig)^{-1}  [\omega ( \bsbI - \bsbSig)  (\bsbb^{(j)} - \bsbb^{(j+1)}) + (1-\omega) (\bsbxi^{(j)} - \bsbxi^{(j+1)})]$.
Furthermore, any limit point $\bsbb^{\circ}$ of $\{\bsbb^{(j)}\}$  is a  group (or non-group) $\Theta$-estimator that  satisfies  \eqref{thetaeq-grp} (or  \eqref{thetaeq}),
and the sequence  $G(\bsbb^{(j)}, \bsbxi^{(j)}, \bsbb^{(j)}, \bsbxi^{(j)};\lambda)$  decreases to the limit $F(\bsbb^{\circ};\lambda)$ with $F$ defined in \eqref{eqn:Grouped Penalized Linear Model} (or \eqref{eqn:Penalized Linear Model}).
\end{theorem}

Applying the theorem to Algorithm \ref{alg:TISP for PLM}, we know the nongroup form solves the optimization problem $\frac{1}{2} \| \bsby - \bsbX \bsbb \|_2^2/\tau_0^2 + \sum_{k=1}^{2D} P(|\beta_k|;\lambda)$, and the group form solves $\frac{1}{2} \| \bsby - \bsbX \bsbb \|_2^2/\tau_0^2 + \sum_{k=1}^{D} P(\sqrt{\beta_k^2 + \beta_{D+k}^2};\lambda)$ for any arbitrarily given $\bsbX$, $\bsby$. Algorithm \ref{alg:TISP for PLM} is  justified for computing a penalized spectrum estimate associated with $P$, provided that  a proper $\Theta$ can be found to satisfy  \eqref{pen-theta}.

The $P$-$\Theta$ strategy covers  most commonly used penalties, either in group form or non-group form. We give some examples below.
(i) When $\Theta$ is the soft-thresholding, the $P$-function according to \eqref{pen-theta} is the  $l_1$-norm penalty used in BP, and the non-group version of our algorithm  reduces to  the iterative soft thresholding    \cite{daubechies2004iterative}. The group $l_1$ penalty  (called the group lasso \cite{Yuan}) is  more suitable for frequency selection, and  can be handled by Algorithm \ref{alg:TISP for PLM} as well.
(ii) When  $\Theta$ is the hard-thresholding, for $q(\cdot;\lambda)\equiv 0$  we get the hard-penalty  \eqref{hard-pen}, and  for $q(t;\lambda)=\frac{(\lambda-|t|)^2}{2} 1_{0 < |t| < \lambda}$ we get the $l_0$-penalty  \eqref{l0-pen}.  The algorithm, in non-group form, corresponds to the iterative hard thresholding  \cite{blumensath2009iterative,blumensath2009normalised}.
(iii) Finally, if we define $\Theta$ to be the hard-ridge thresholding:
\begin{equation} \label{eqn:hrth}
\Theta_{HR}(t; \lambda,\eta)= \begin{cases}
0, & \text{if }|t|<\lambda \\
{t\over 1+\eta}, & \text{if }|t|\geq\lambda. \\
\end{cases}
\end{equation}
then $P_{\Theta_{HR}}$ is the hard-ridge penalty \eqref{hrpen}. Setting $q(t;\lambda, \eta)=\frac{1+\eta}{2} (|t|-\lambda)^2  1_{0 < |t| < \lambda}$, we successfully reach the $l_0+l_2$ penalty  \eqref{l02-pen}. See \cite{SheGLMTISP} for more examples, such as  SCAD, $l_p$ ($0< p<1$), and elastic net.

Algorithm \ref{alg:TISP for PLM} includes a relaxation parameter $\omega$, which is an effective means to accelerate the convergence. See the recent work by   Maleki \&  Donoho \cite{maleki:optimally}. (Our relaxation form is novel and is of  Type I based on  \cite{she2010spareg}).
In practice, we  set $\omega=2$, and the number of iterations can be reduced by about 40\% in comparison to nonrelaxation form.

\subsection{Statistical analysis}
\label{subsec:statana}

Although the $l_1$ regularization  is popular (see, e.g.,  BP \cite{chen1998application}), in the following  we show that the HR penalty has better selection power and can remove the stringent coherence assumption and can accommodate lower SNRs. We focus on the group form based on the  discussion in Section \ref{sec:Model}.

Let $\mathcal F$ be the entire frequency set  covered by the dictionary. For the design matrix  defined in \eqref{eqn:X0}, $\mathcal F=\{f_1, \cdots, f_D\}$.  Given any frequency $f\in \mathcal F$, we use $\bsbX_f$ to denote the submatrix of $\bsbX$ formed by the sine and cosine frequency atoms at $f$, and $\bsbb_f$ the corresponding coefficient vector. If $I\subset \mathcal F$ is  an index set, $\bsbX_I$ and $\bsbb_I$ are defined similarly. In general, $\bsbX_f$ is of size $N\times 2$ and $\bsbb_f$ $2\times 1$ (but not always--cf. \eqref{eqn:X}).
Given any coefficient vector $\bsbb$, we introduce
\begin{align}
z(\bsbb) = \{f\in \mathcal F: \|\bsbb_f\|_2 = 0\}, nz(\bsbb) = \{f\in \mathcal F: \|\bsbb_f\|_2   \neq 0\} \label{znzdef}
\end{align}
  to characterize the frequency selection outcome.
In particular, we write  $z^*=z(\bsbb^*)$, $nz^*=nz(\bsbb^*)$, associated with  the true coefficient vector $\bsbb^*$, and let $p_{nz^*}=|nz^*|$ be the number of frequencies present in the true signal, and $p_{z^*}=|z^*|$  the number of irrelevant frequencies.



We introduce two useful  quantities $\kappa$ and $\mu$.
Recall  $\bsbSig=\bsbX^T \bsbX$ and $\tau_0^2 = \| \bsbSig \|_2 = \mu_{\max}(\bsbSig)$ (the largest eigenvalue of $\bsbSig$).  Given  $I\subset \mathcal F$, let $\bsbSig_{{{I}, {I'}}}=\bsbX_{ I}^T \bsbX_{{I'}}$ and $\bsbSig_{{{I}}}=\bsbX_{ I}^T \bsbX_{{I}}$.
In this subsection, we assume the design matrix has been column-normalized such that the 2-norm of every column is $\sqrt N$. Let $\bsbSig^{(s)}=\bsbSig/N$. Define
\begin{align*}
\mu :=  \mu_{\min}(\bsbSig_{nz^*,nz^*}^{(s)}), \mbox{ and }
\kappa :=  \underset{f \in z^*}{\max} \| \bsbSig_{f,nz^*}^{(s)} \|_2 / \sqrt{ p_{nz^*}},
\end{align*}
where $\mu_{\min}$ denotes the smallest eigenvalue and $\|\cdot\|_2$ refers to  the spectral norm. ($\bsbSig_{f,nz^*}^{(s)}$ is of size $2\times  2p_{nz^*}$  typically.) Intuitively, $\kappa$ measures the `mean' correlation between the relevant frequency atoms and the irrelevant atoms.
 When $\kappa$ is high, the coherence of the dictionary is necessarily high.
Denote by  $P_1$  the probability that with soft-thresholding  being applied, there exists at least one estimate  $\hat \bsbb$ from \textbf{Algorithm} \ref{alg:TISP for PLM} such that $nz(\hat\bsbb) = nz^*$.  $P_{02}$ is similarly defined  for hard-ridge thresholding.
Theorem \ref{th:sel} bounds these two probabilities.

\begin{theorem}\label{th:sel}
Assume $\mu>0$.  \\
(i) Let  $\Theta$ be the soft-thresholding.
Under the assumption that  $\kappa < \mu / p_{nz}$ and $\lambda$ is chosen such that $\min_{f\in nz^*}\|\bsbb_{f}^*\|_2 \geq \frac{\lambda \sqrt{p_{nz^*}}}{N\mu/\tau_0^2 }$, we have
\begin{align}
1-P_1\leq \frac{e}{4} \left( \frac{p_{z^*} M^2}{ e^{M^2/4}} +  \frac{p_{nz^*} L^2}{ e^{L^2/4}}\right), \label{selbndL1}
\end{align}
where $M:=\frac{\lambda\tau_0^2}{\sigma\sqrt N}(1-\frac{\kappa p_{nz^*}}{\mu})$ and $L:=(\min_{f\in nz^*}\|\bsbb_{f}^*\|_2 - \frac{\lambda\tau_0^2 \sqrt{p_{nz^*}}}{N\mu })\frac{\sqrt{ N\mu}}{\sigma}$.
\\
(ii) Let  $\Theta$ be the hard-ridge thresholding. Assume  $\lambda,\eta$ are chosen such that $\kappa \leq \frac{1}{\eta}\frac{\lambda (\mu N+\eta \tau_0^2)}{\|\bsbb_{nz^*}^*\|_2\sqrt{ p_{nz^*}}}$, $\iota:=\min_{f\in nz^*} \|[(\bsbSig_{nz^*}+\eta \bsbI)^{-1} \bsbSig_{nz^*} \bsbb_{nz^*}^*]_f\|_2 \geq \frac{\lambda}{1+\eta}$, and $\eta \leq \mu N/\tau_0^2$. Then
\begin{align}
1-P_{02} \leq  \frac{e}{4} \left( \frac{p_{z^*} M'^2}{ e^{M'^2/4}}+   \frac{p_{nz^*} L'^2}{ e^{L'^2/4}}\right), \label{selbndL02}
\end{align}
where $M':= \frac{1}{\sigma\sqrt N}(\lambda \tau_0^2- \frac{\eta \tau_0^2}{ \mu N+\eta \tau_0^2} \kappa \sqrt{p_{nz^*}} \|\bsbb_{nz^*}^*\|_2)$ and $L':=(\iota- \frac{\lambda}{1 + \eta})\frac{\sqrt{\mu N}+ \eta\tau_0^2/\sqrt{\mu N}}{\sigma}$.
\end{theorem}



Seen from \eqref{selbndL1} and \eqref{selbndL02}, both  inconsistent detection probabilities  are small. It is worth mentioning that in practice, we found the value of $\eta$ is usually  small, which, however,   effectively handles singularity/collinearity in comparison to $\eta=0$, as  supported by the literature (e.g.,    \cite{Park07}).
In the following,   we make a comparison of the assumptions and  probability bounds. The setup of $p_{z^*}\gg N \gg p_{nz^*}$ is of particular interest, which means the number of truly present frequencies is small relative to the sample size but the number of irrelevant frequencies is overwhelmingly large.
The $\kappa$-conditions  characterize coherence accommodation, while the conditions on $\min_{f\in nz^*}\|\bsbb_{f}^*\|_2$ and $\iota$ describe  how small the minimum signal strength can be.
(i) For the $l_1$ penalty, $\kappa < \mu/p_{nz^*}$ is a version of the irrepresentable conditions  and cannot be relaxed in general \cite{zhao2006model}. In contrast, for the $l_0+l_2$, the bound for $\kappa$ becomes large when $\eta$ is small, and so the stringent coherence requirement can be essentially removed!
(ii)
When $\eta$ is small  in the hard-ridge thresholding, the noiseless ridge estimator $(\bsbSig_{nz^*}+\eta \bsbI)^{-1} \bsbSig_{nz^*} \bsbb_{nz^*}^*$  is close to $\bsbb_{nz^*}^*$, but the minimum signal strength  can be much lower than that of the $l_1$, due to the fact that  $N \mu/\tau_0^2 =   \mu_{\min}(\bsbSig_{nz^*,nz^*}^{(s)})/\mu_{\max}(\bsbSig^{(s)})  \leq 1\leq  1+\eta$ and in particular, the disappearance of $\sqrt{ p_{nz^*}}$.
(iii)
Finally, for small values of $\eta$, $M'>M$, $L'>L$, and so $l_0+l_2$ has a better chance to recover the whole spectra correctly.

\emph{Remark.} Including the  ridge penalty in regularization is helpful to enhance  estimation and prediction  accuracy, especially when the frequency resolution is quite high and the true signal is multi-dimensional. Even when the purpose is  selection alone, it is  meaningful because most  tuning strategies of $\lambda$ are prediction error (generalization error) based.


\subsection{Model comparison criterion}
\label{subsec:Selection}
This part studies the  problem of how to choose proper  regularization parameters  for any given data $(\bsbX, \bsby)$. In \eqref{eqn:Grouped Penalized Linear Model}, the general parameter $\lambda$ provides a statistical bias-variance tradeoff in regularizing the model, and  ought to   be tuned in a data-driven manner.
In common with most researchers (say \cite{sparspec,chenchen,SLIM}),  we first specify a grid $\Lambda=\{\lambda_1,\cdots,\lambda_l,\cdots,\lambda_L\}$, then run Algorithm
\ref{alg:TISP for PLM} for every $\lambda$ in the grid to get a \emph{solution path} $\hat \bsbb(\lambda_l)$, $1\leq l \leq L$, and finally, use a \emph{model comparison criterion} to find the optimal estimate $\hat \bsbb_{opt}$. The commonly used model comparison criteria are  Akaike information criterion  (AIC), Bayesian information criterion (BIC), and cross-validation (CV). But we found none of them is satisfactory in the high-dimensional super-resolution spectral estimation.

Ideally, in a data-rich situation, one would  divide the whole dataset into a training subset denoted by $(\bsbX^{trn}, \bsby^{trn})$ and a validation subset $(\bsbX^{val}, \bsby^{val})$. For any $\lambda\in \Lambda$, train the model  on $(\bsbX^{trn}, \bsby^{trn})$ and  evaluate the prediction accuracy on the validation subset by, say,   $\|\bsby^{val} - \bsbX^{val} \hat\bsbb(\lambda)\|_2^2$. 
However, this data-splitting approach is only reasonable when the validation subset is large enough to approximate the true  prediction error.  It cannot be used in our problem due to insufficiency of observations.
A popular data-reusing method in small samples  is the $\mathpzc K$-fold  CV.
Divide the dataset into $\mathpzc K$ folds. Let $(\bsbX^{({\mathpzc k})}, \bsby^{(\mathpzc{k})})$ denote  the $\mathpzc k$th subset, and $(\bsbX^{(-{\mathpzc k})}, \bsby^{(-\mathpzc{k})})$ denote  the remaining data.  To obtain the CV error at any $\lambda_l\in \Lambda$, one needs to fit $\mathpzc K$ penalized models. Concretely, setting $\bsbX=\bsbX^{(-\mathpzc k)}$ and $\bsby=\bsby^{(-\mathpzc k)}$ as the training data, solve the penalized problem associated with $\lambda_l$, the estimate represented by $\hat\bsbb^{(-\mathpzc k)}(\lambda_l)$. Then calculate the validation error on $(\bsbX^{({\mathpzc k})}, \bsby^{(\mathpzc{k})})$: $\mbox{cv-err}(\lambda_l, \mathpzc k)=\|\bsby^{(\mathpzc{k})}- \bsbX^{(\mathpzc{k})} \hat\bsbb^{(-\mathpzc k)}(\lambda_l)\|_2^2$. The summarized CV error,   $\mbox{cv-err}(\lambda_l)=\sum_{\mathpzc k = 1}^{\mathpzc K} \mbox{cv-err}(\lambda_l, \mathpzc k)/N$, serves as the comparison criterion. After the optimal $\lambda_{opt}$ is determined, we refit the model on the global dataset to get $\hat\bsbb_{opt}$.

\begin{figure}[h!]
\centering
\includegraphics[width=3.5in,height=3in]{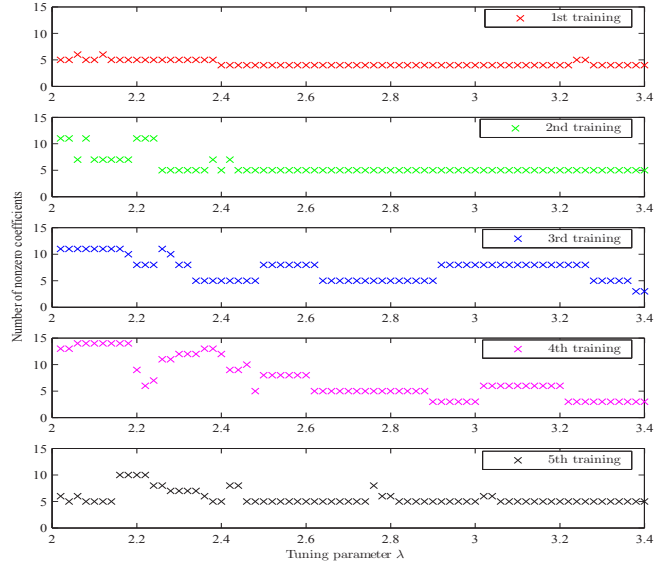}
\caption[]{{The numbers of nonzero coefficients in $5$-fold CV with respect to $\lambda$. The $5$ CV trainings  yield (sometimes quite) different models at the same  value of $\lambda$.} }
\label{fig:lambdapath}
\end{figure}

\begin{figure}[h!]
\vspace{0cm}
\centering
\includegraphics[width=3.5in,height=3in]{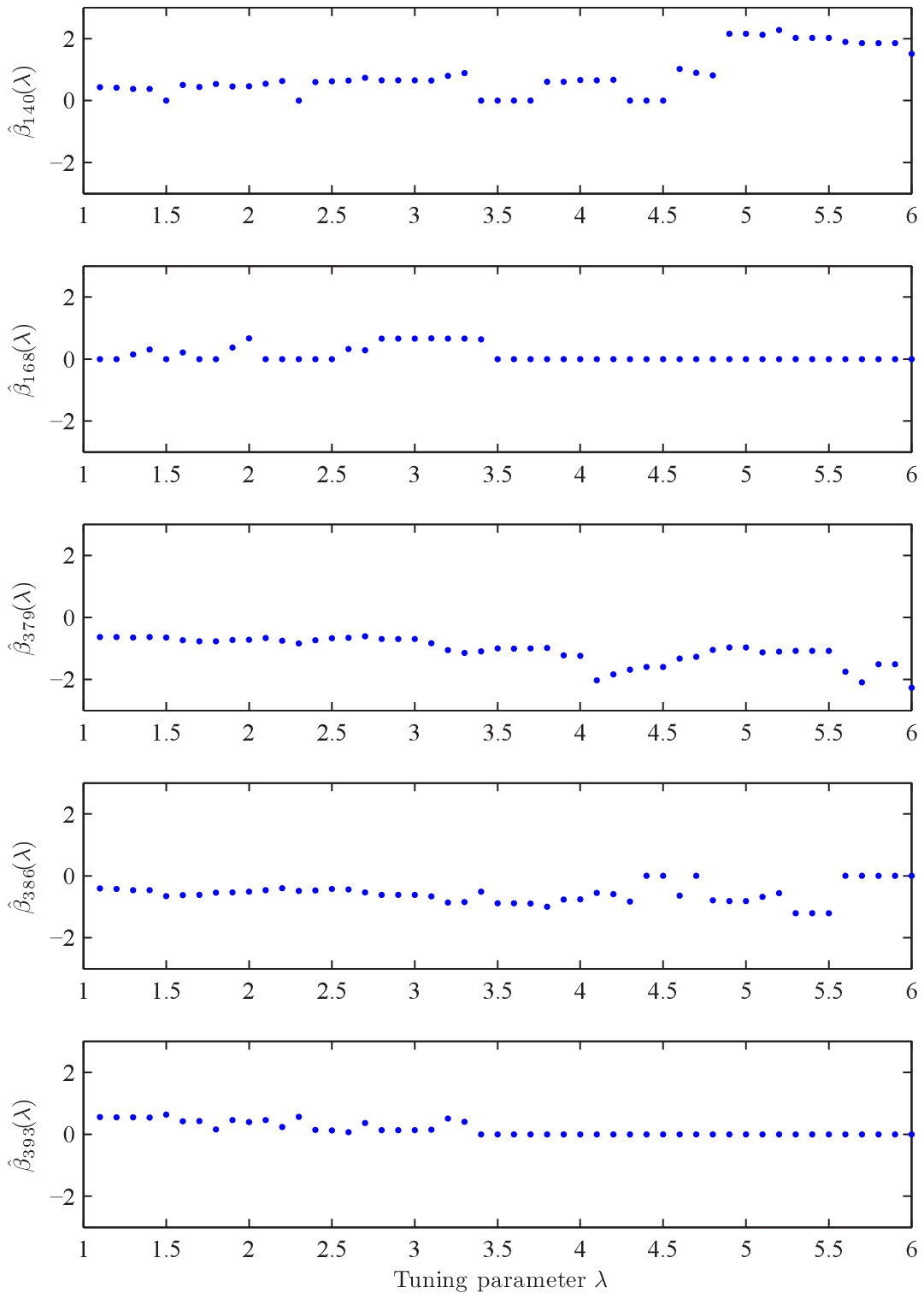}
\caption[]{{The $l_0$-penalized solution path  $\hat\bsbb(\lambda)$ is discontinuous in $\lambda$.   
For clarity, only 5 frequency paths (chosen at random) are shown.}
}
\label{fig:solpath}
\end{figure}

However,  when a nonconvex penalty is applied, the above plain CV has an inherent drawback:  the $\mathpzc K$  trained models at a common value of $\lambda_l$ may  not  be comparable, and thus  averaging their validation errors   may make little sense. The reasons are twofold.
(i) The regularization parameter $\lambda$ appears in a Lagrangian form optimization problem (cf. \eqref{eqn:Penalized Linear Model} or \eqref{eqn:Grouped Penalized Linear Model}). In general, the optimal $\lambda$ to guarantee good selection and estimation must be a function of both the true coefficient vector $\bsbb^*$ and the data $(\bsbX,\bsby)$.  Notice that in the trainings of $\mathpzc K$-fold  CV, $(\bsbX, \bsby)$ changes. The same value of $\lambda$ may have different regularization effects  for  different training datasets  although   $\bsbb^*$ remains the same.
 Fig.~\ref{fig:lambdapath} shows the numbers of nonzero coefficient estimates under the $l_0$ penalization in $5$-fold CV---they are never consistent at any fixed value of $\lambda$!
(ii) The solution path $\hat \bsbb(\lambda)$ associated with a nonconvex penalty is generally discontinuous in $\lambda$.
Fig.~\ref{fig:solpath} plots the $l_0$ solution path for the default TwinSine signal.
Even a small change in  $\lambda$ may result in a totally different estimate and zero-nonzero pattern.
In consideration of both (i) and (ii),   cross-validating $\lambda$ is not a proper tuning strategy in our problem.


To resolve the training inconsistency, we advocate a generic selective  cross validation (\textbf{SCV}) for parameter tuning in sparsity-inducing penalties. First the sparsity algorithm is run on the \emph{entire} dataset to get a solution path  $\hat\bsbb(\lambda_l)$, $l=1,\cdots, L$. Every estimate $\hat\bsbb(\lambda_l)$ determines a candidate model with the predictor set given by $nz_l=nz(\hat\bsbb(\lambda_l))=\{f_k\in \mathcal{F}: \hat\beta_k^2+\hat\beta_{k+D}^2 \neq 0\}$. Next, we \emph{cross-validate $nz_l$} (instead of $\lambda$) to evaluate the goodness-of-fit of each candidate model. In this way, all $\mathpzc K$ trainings are   restricted to the \emph{same} subset of predictors.
Concretely, for  penalties without $l_2$ shrinkage, such as the $l_0$-penalty, $\hat \bsbb^{(-\mathpzc{k})}(\lambda_l)$ is the unpenalized regression estimate fitted on $(\bsby^{(-\mathpzc{k})},  \bsbX_{nz_l}^{(-\mathpzc{k})})$, while for penalties with $l_2$ shrinkage, such as the $l_0+l_2$-penalty, $\hat \bsbb^{(-\mathpzc{k})}(\lambda_l)$ is the ridge regression estimate fitted on $(\bsby^{(-\mathpzc{k})},  \bsbX_{nz_l}^{(-\mathpzc{k})})$ (cf.   Theorem \ref{th:conv}), i.e., $\hat \bsbb^{(-\mathpzc{k})}(\lambda_l)=((\bsbX_{nz_l}^{(-\mathpzc{k})})^T \bsbX_{nz_l}^{(-\mathpzc{k})} + \eta \bsb{I})^T (\bsbX_{nz_l}^{(-\mathpzc{k}) })^T \bsby^{(-\mathpzc{k})}$. Finally, the total SCV error is summarized by $\mbox{SCV}(\lambda_l) = \sum_{\mathpzc k = 1}^{\mathpzc K}\|\bsby^{(\mathpzc{k})}- \bsbX^{(\mathpzc{k})} \hat\bsbb^{(-\mathpzc k)}(\lambda_l)\|_2^2$.

Motivated by the work of  \cite{chenchen}, we add a high-dimensional BIC correction term to define the model comparison criterion: $\mbox{SCV-BIC}(\lambda_l) =  \mbox{SCV}(\lambda_l)+   \mbox{DF}(\hat\bsbb(\lambda_l)) \log N$, where   $\mbox{DF}$ is the degrees  of freedom function.
When the true signal has a parsimonious representation in the frequency domain, i.e.,  the number of present frequencies is very small, such a  correction is necessary---see \cite{chenchen}  for a further theoretical justification.
For the $l_0$ or $l_1$ penalty, $\mbox{DF}$ is approximately the number of nonzero components in the  estimate; for the $l_0+l_2$ penalty,  $\mbox{DF}(\hat\bsbb(\lambda_l))$ is given by $Tr((\bsbX_{nz_l}^T\bsbX_{nz_l}+\eta \bsbI)^{-1} \bsbX_{nz_l}^T\bsbX_{nz_l})$ \cite{Park07}.
The  optimal estimate $\hat\bsbb_{opt}$ is chosen from the original solution path $\{\hat\bsbb(\lambda_l)\}_{l=1}^L$ by minimizing $\mbox{SCV-BIC}(\lambda_l)$.

We point out  that in SCV, the sparsity algorithm is only required to run  on the whole dataset to generate one solution path, while CV needs   $\mathpzc K$ such solution paths. SCV is  more efficient in computation.

\subsection{Probabilistic spectra screening}
\label{sec:Screening}
Computational complexity is another major challenge in super-resolution studies.
In Algorithm \ref{alg:TISP for PLM}, each iteration step  involves only matrix-vector multiplications and componentwise thresholding operations. Both have low complexity and can be vectorized. The total number of flops  is no more than  $(4DN + 8D)\Omega$,   which is linear in $D$. In our experiments, $\Omega=200$ suffices and thus the complexity of Algorithm  \ref{alg:TISP for PLM} is $O(DN)$. (Restricting  attention to uniformly sampled data and frequency atoms in the dictionary construction, we can use the Fast Fourier transform (FFT) in computation to reduce the complexity to $O(D \log D)$, as pointed out by an anonymous reviewer, see \cite{SLIM-cg} and Section \ref{sec:Experiments}.)
On  the other hand, with a superbly high resolution dictionary (where  $D$ is very large), dimension reduction is still desirable to further reduce the computational cost.

%

This is indeed possible  under the spectral sparsity assumption, where the number of true components is supposed to be much smaller than $N$.  One may reduce the dimension from $2D$ to $\vartheta N$ (say $\vartheta=0.5$)  before running the formal algorithm. If  the $\vartheta N$ candidate predictors are wisely chosen, the truly relevant atoms will be included with high probability and the  performance sacrifice in selection/estimation will be mild. Hereinafter, we call  $\vartheta$  the \emph{candidate ratio}. A well designed screening algorithm  should not be very sensitive to $\vartheta$ as long as it is reasonably large. Significant decrease in computational time can be achieved after this supervised dimension reduction.

We propose an {iterative probabilistic  screening}  by adapting
  Algorithm \ref{alg:TISP for PLM}  for dimension reduction.
 This has the benefit that the screening principle is  consistent with the fitting criterion.
We recommend using the hard-ridge thresholding  and the associated Algorithm \ref{alg:GIST-screening} is stated below.


\begin{algorithm}
\begin{algorithmic}
\caption{GIST-Screening algorithm. \label{alg:GIST-screening}}
\STATE \given\ $\bsbX$ (design matrix, normalized), $\bsby$ (centered),
$\eta$ ($l_2$ shrinkage parameter),
 $\vartheta$ (candidate ratio--ratio of new dimension to  sample size),
 $\omega$ (relaxation parameter),
and $\tilde\Omega$ (maximum number of iterations).
(For simplicity, assume $\vartheta N$ is an integer.)

\STATE 1) $\bsbX\leftarrow \bsbX/\tau_0$, $\bsby\leftarrow \bsby/\tau_0$, 
with  $\tau_0\geq \|\bsbX\|_2$ (spectral norm).
\STATE  2) Let  $j\leftarrow 0$ and $\bsbb^{(0)}$ be an initial estimate say $\bsb{0}$.
\WHILE{$\|\bsbb^{(j+1)}-\bsbb^{(j)}\|$ is not small enough or $j \leq \tilde \Omega$}

\STATE 3.1)
   $\bsbxi^{(j+1)} \leftarrow (1-\omega)\bsbxi^{(j)}+\omega ( \bsbb^{(j)}+ \bsbX^T(\bsby-\bsbX\bsbb^{(j)}) )$ if $j>0$ and
  $\bsbxi^{(j+1)} \leftarrow    \bsbb^{(j)}+ \bsbX^T(\bsby-\bsbX\bsbb^{(j)})$ if $j=0$, and set $m^{(j)} = \vartheta N$; \\
 \underline{{\sc Group form}:}\\
\STATE 3.2a)  $l_k^{(j+1)}\leftarrow \sqrt{ (\xi_{k}^{(j+1)} ) ^2 + (\xi_{k+D}^{(j+1)}) ^2}$, $1\leq k \leq D$
\STATE 3.2b)  \emph{Let $\lambda$ be the median of the $m^{(j)}$th largest and ($m^{(j)}+1$)th largest
elements in $\{\bsb{l}_k^{(j+1)}\}$.}
  For each $k:1\leq k \leq D$,
  if $l_k^{(j+1)}\neq 0$,
  set $[\bsbb_{k}^{(j+1)}, \bsbb_{k+D}^{(j+1)}] \leftarrow [{\xi_{k}^{(j+1)} }, {\xi_{k+D}^{(j+1)}  }] \Theta_{HR}( l_k^{(j+1)};{\lambda}, \eta )/ l_k^{(j+1)}$;  set $\bsbb_{k}^{(j+1)}=\bsbb_{k+D}^{(j+1)}=0$ otherwise.

\STATE \underline{{\sc Non-Group form}:}\\
\STATE 3.2')     \emph{$\bsbb^{(j+1)}\leftarrow \Theta_{HR}( \bsbxi^{(j+1)}; {\lambda}, \eta)$, where   $\lambda$ is the median of the $m^{(j)}$th}
 \emph{largest component and the $(m^{(j)}+1)$th largest component  of $|\bsbb^{(j+1)}|$};
\ENDWHILE
\STATE\deliver\   Remaining dimensions after screening:
 $\{f\in \mathcal F:  \| \bsbb_f^{(j+1)}\|_2 \neq 0 \}$  (group version)
 or $\{k:1\leq k \leq 2D, \beta_k^{(j+1)} \neq 0 \}$ (non-group version).
\end{algorithmic}
\end{algorithm}

The differences in comparison to   Algorithm \ref{alg:TISP for PLM} lie  in (3.2b) and (3.2'), where a dynamic threshold is constructed in performing the hard-ridge thresholding.
We next show that this screening version still has convergence guarantee.  Similar to Theorem \ref{th:conv}, assume $\tau_0 = 1 > \|\bsbX\|_2$. Let $G$ be the same energy function constructed in \eqref{defG} with  $P$ given by \eqref{hrpen} or  \eqref{l02-pen}. For simplicity, suppose $m:=\vartheta N\in \mathbb N$. Theorem \ref{th:scr} shows that Algorithm \ref{alg:GIST-screening} solves an $l_0$-constrained problem.
\begin{theorem}\label{th:scr}
For any  $0<\omega\leq 1$,  the sequence of iterates $(\bsbb^{(j)},\bsbxi^{(j)})$ from Algorithm \ref{alg:GIST-screening} has the same function value decreasing property \eqref{fundecrease}  for the energy function $G$, and $\bsbb^{(j)}$ satisfies  $nz(\bsbb^{(j)})\leq m$.
In addition, under $\eta>0$ and the no-tie-occurring assumption $\mathfrak B$ in Appendix \ref{app:proofscrconv},
the sequence of  $\bsbb^{(j)}$  has a unique limit point  $\bsbb^{\circ}$ which corresponds to  the ridge estimate restricted to $\bsbX_{nz(\bsbb^{\circ})}$ with $|nz(\bsbb^{\circ})| \leq m$.
\end{theorem}

We can use SCV to tune $\eta$ or simply set $\eta$ at a small value (say 1e-2).
In practice, the screening  can proceed in a progressive fashion to  avoid greedy selection:  we use a varying sequence of $m^{(j)}$ that decreases to $\vartheta N$  in Step 3.1),  and add  `squeezing' operations after Step 3.2) or 3.2'): ${\bsb{d}} \gets \{f\in \mathcal F: \|\bsbb_f^{(j+1)}\|_2 \neq 0\},  \bsbb^{(j+1)} \gets \bsbb^{(j+1)}[{\bsb{d}}], \bsbX \gets \bsbX[, {\bsb{d}}]$ (group version),  or   ${\bsb{d}} \gets \{k: 1\leq k \leq 2D,  \bsbb_k^{(j+1)} \neq 0\}, \bsbb^{(j+1)} \gets \bsbb^{(j+1)}[{\bsb{d}}], \bsbX \gets \bsbX[, {\bsb{d}}]$ (non-group version).
We have  found that empirically, the sigmoidal decay cooling schedule
$
m^{(j)} =
\lceil {2 D}/(1+\exp(\alpha j))\rceil
$
with $\alpha=0.01$ achieves good balance between selection and efficiency.

GIST-Screening  works decently in super-resolution spectral analysis seen from the experiments:
after dimension reduction the true signal components are  included  with high probability and  the  computational cost can be significantly reduced.

An interesting observation is that with $\bsbb^{(0)}=\bsb{0}$, the first iteration step of Algorithm \ref{alg:GIST-screening}  ranks the frequencies based on $\bsbX^T \bsby$. In other words,  the correlation between the signal $\bsby$ and each dictionary atom is examined separately, to determine the candidate dimensions, see \cite{fan2008sure}. Of course, this type of single frequency analysis is merely marginal and does not amount to joint  modeling, the resulting  crude ranking  not suitable for super resolution problems due to the existence of many correlated frequency predictors.
Algorithm \ref{alg:GIST-screening} iterates  and avoids such greediness.

GIST screening is pretty flexible and useful, even if sparsity is not desired. It  can be applied at any given value of  $\vartheta$ (possibly greater than 1) and  yields a meaningful result for super-resolution spectral problems.

\subsection{GIST framework}
\label{subsec:Flowchart}
\begin{figure}[h!]
  \centering
  \includegraphics[width=0.5\hsize, height=2.5in]{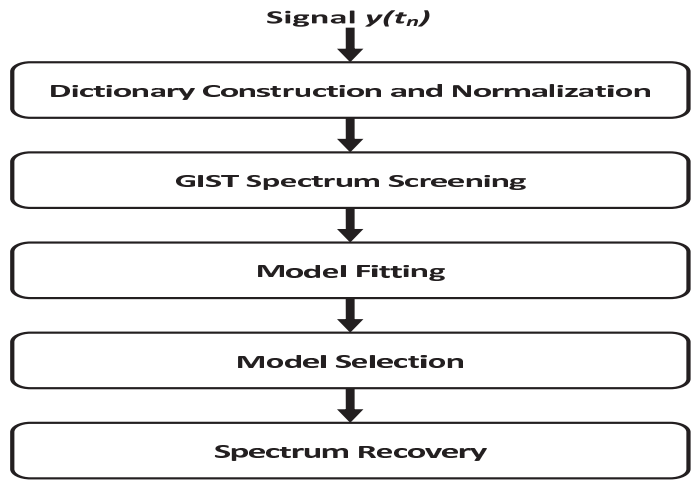}
  \caption{\small{The flowchart of the GIST framework  for solving the spectral estimation problem.}}
  \label{fig:flowchart}
\end{figure}

We introduce the complete GIST framework to solve the spectral estimation problem.
Fig.~\ref{fig:flowchart} shows the flowchart outline.

\begin{enumerate}
\item \emph{Dictionary Construction and Normalization}:
We construct an overcomplete dictionary through \eqref{eqn:X0} with sufficiently high resolution.
Then standardize the data, by (a) centering $\bsby$ and (b) normalizing each predictor column in $\bsbX$ to have mean 0 and variance 1. 
After the standardization,  all predictors are equally extended in the predictor space.

\item \emph{GIST Spectrum Screening}:
This  step can greatly reduce the computational complexity. 
We perform the iterative probabilistic screening to remove a number of  nuisance frequency components and keep $\vartheta N$ candidate predictors with $\vartheta<1$ (say $\vartheta=0.5$) to achieve supervised dimension reduction. See  Section \ref{sec:Screening} for details.

\item \emph{Model Fitting}:
For each given value of the regularization parameter in a predefined grid,  
run the iterative group-thresholding algorithm developed in Section \ref{subsec:Solving} to obtain a local optimum to  \eqref{eqn:Grouped Penalized Linear Model}. All such solutions are collected to form a solution path parameterized by the regularization parameters.

\item \emph{Model Selection}: An optimal solution $\hat{\bsbb}_{opt}$ is selected from the solution path based on a data-resampling version of high-dimensional BIC  (Section~\ref{subsec:Selection}).

\item \emph{Spectrum Recovery}:
The signal can be reconstructed from the coefficient estimate.
The  amplitudes are estimated by $A(f_k)=\sqrt{\hat{\bsbb}_{opt,k}^2+\hat{\bsbb}_{opt,D+k}^2}$, $1\leq k \leq D$.
\end{enumerate}

\section{Experiments}
\label{sec:Experiments}

We conduct simulation  experiments to show the performance of GIST fitting Algorithm \ref{alg:TISP for PLM} in sparse spectral estimation, and the power of GIST screening Algorithm \ref{alg:GIST-screening} in fast computation (with little performance loss in frequency detection).


\subsection{Simulation setup}
\label{subsec:data}

Consider a discrete real-valued  signal given by
\begin{align}
y(t_n)= \sum_{f_k \in nz^*} A_k \cos(2\pi f_k t_n+\phi_k)+e(t_n),
\end{align}
where  $e(t_n)$ is white Gaussian noise with variance $\sigma^2$.
$N=100$ training samples are observed at time $t_n=n$, $1 \leq n \leq N$.
The spectrum  frequency dictionary is constructed by setting the maximum frequency $f_{\max}=0.5$ Hz,  resolution level $\delta=0.02Hz$, and  the number of frequency bins   $D= f_{\max} / \delta=250$ (and thus $500$ atoms). Using the notation in Section \ref{subsec:statana} (cf. \eqref{znzdef}), we set $nz^* = \{ 0.248, 0.25, 0.252,  0.398, 0.4\}$, the associated amplitudes $A_k$ and phases $\phi_k$ given by $[2, 4, 3, 3.5, 3]$ and $[\pi/4, \pi/6, \pi/3, \pi/5, \pi/2]$, respectively.
We   vary the noise level by $\sigma^2 =1, 4, 8$ to study the algorithmic performance with respect to SNR.

Due to random fluctuation,   reporting frequency identification for \emph{one} particular simulation dataset is {meaningless}.
Instead, we simulated each model  50 times to enhance stability, where at each run $e(t_n)$ are i.i.d. following $\mathcal N(0, \sigma^2)$.

Our simulations were performed in MATLAB R2010b and Win7 Professional 32-bit OS,
on a desktop with an Intel(R) Core(TM)2 Quad 2.66 GHz processor and 4GB
memory.

\subsection{Experimental Results}


\subsubsection{Comparison with some existing methods}
\label{subsec:exp sparse}
\begin{figure*}[h!]
\centering

\begin{subfigure}[b]{0.4\textwidth}
    \centering\includegraphics[width=\textwidth]{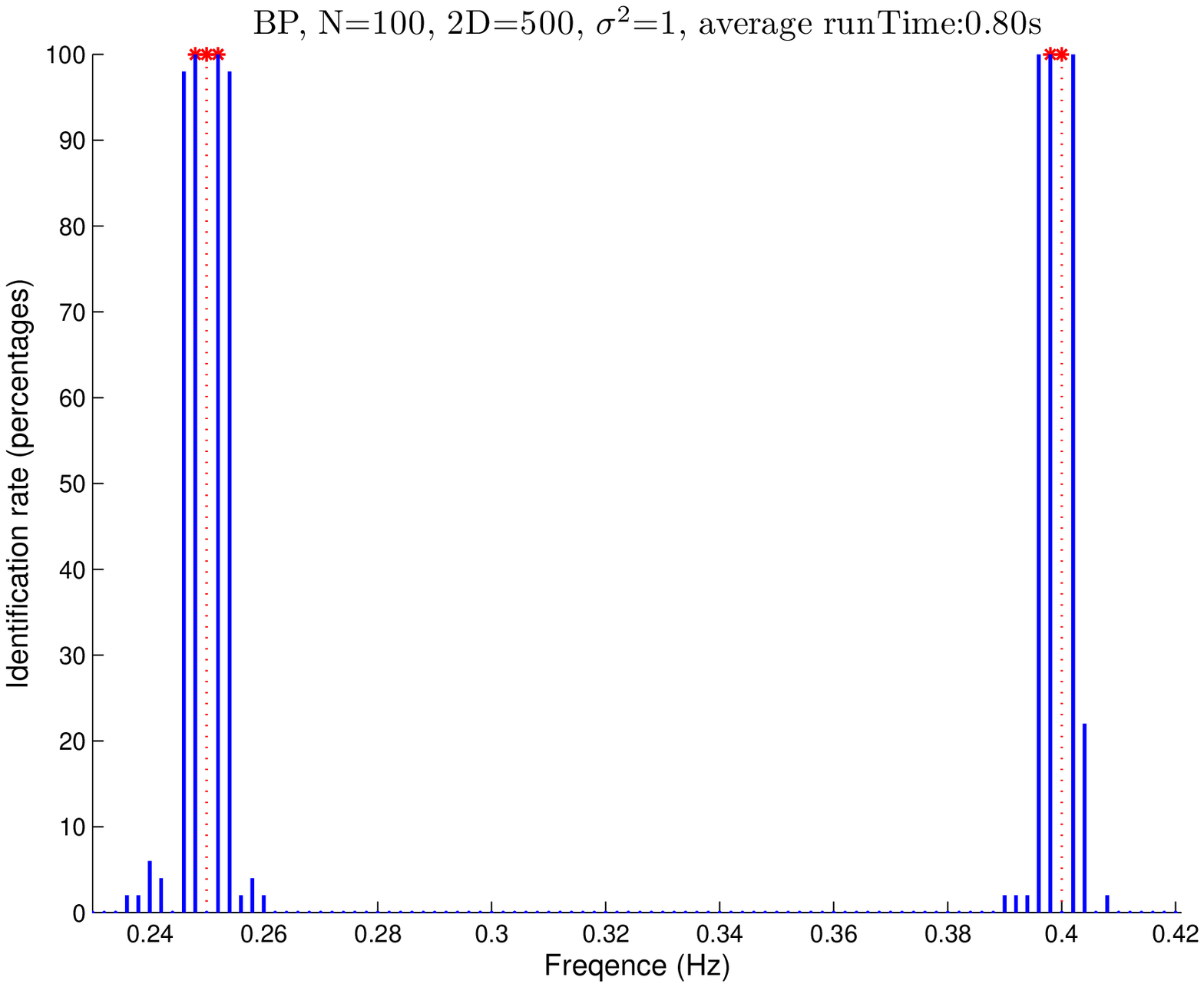}
    \caption{BP}
            \label{subfig:BP_iden1}
\end{subfigure}\hfill
\begin{subfigure}[b]{0.4\textwidth}
    \centering\includegraphics[width=\textwidth]{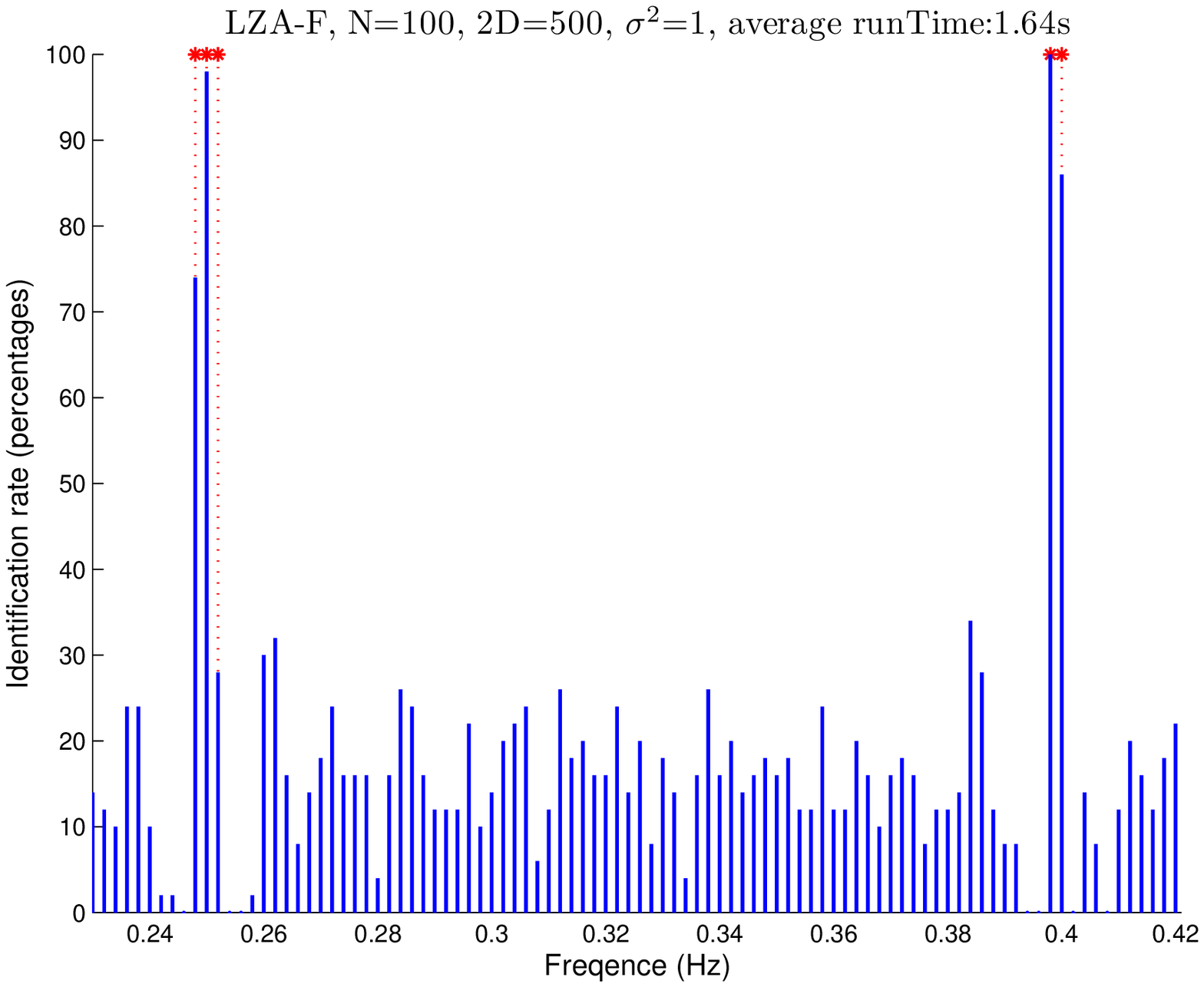}
    \caption{LZA-F}
                    \label{subfig:LZAF_iden1}
\end{subfigure}

\begin{subfigure}[b]{0.4\textwidth}
    \centering\includegraphics[width=\textwidth]{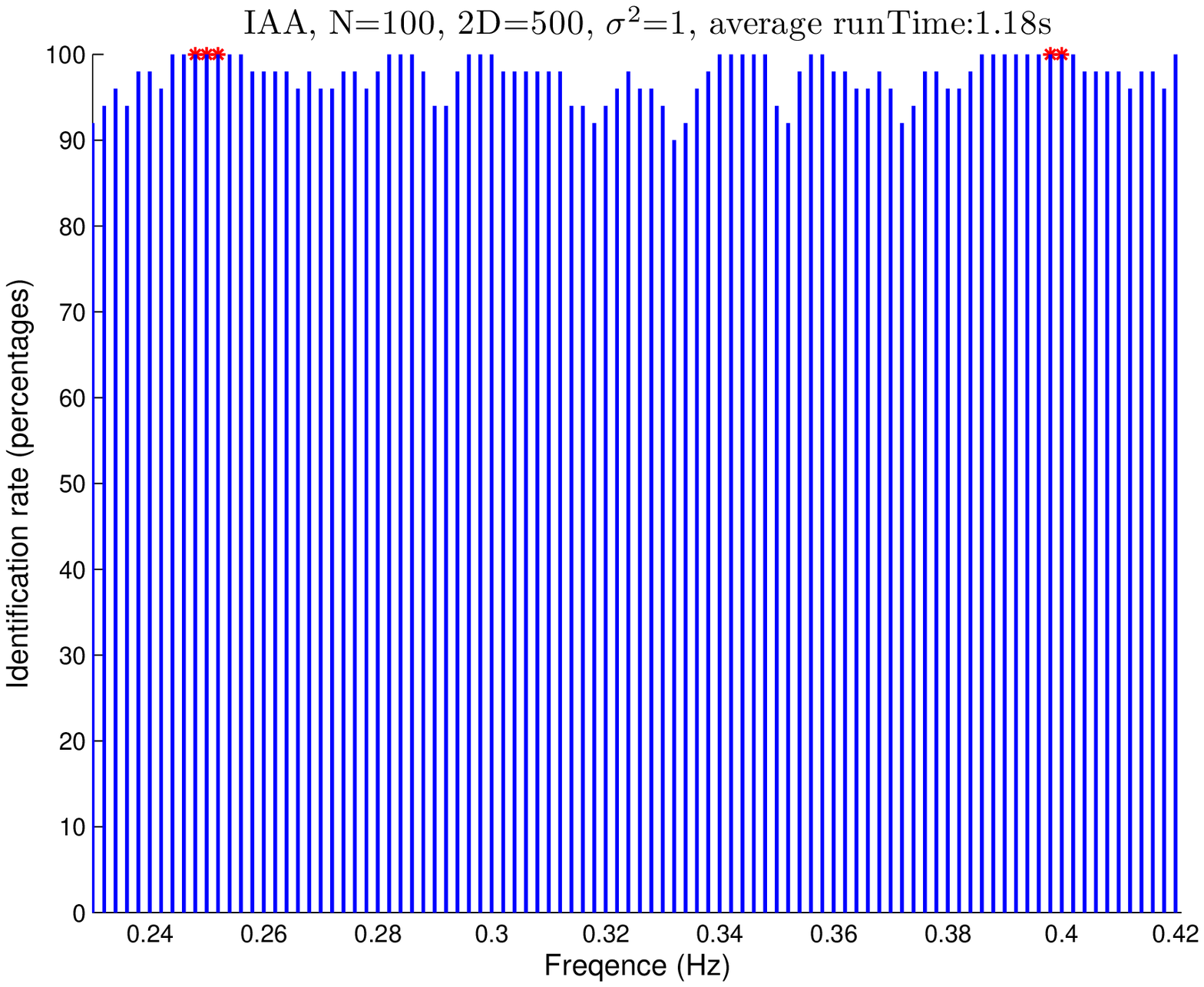}
    \caption{IAA}
            \label{subfig:IAA_iden1}
\end{subfigure}\hfill
\begin{subfigure}[b]{0.4\textwidth}
    \centering\includegraphics[width=\textwidth]{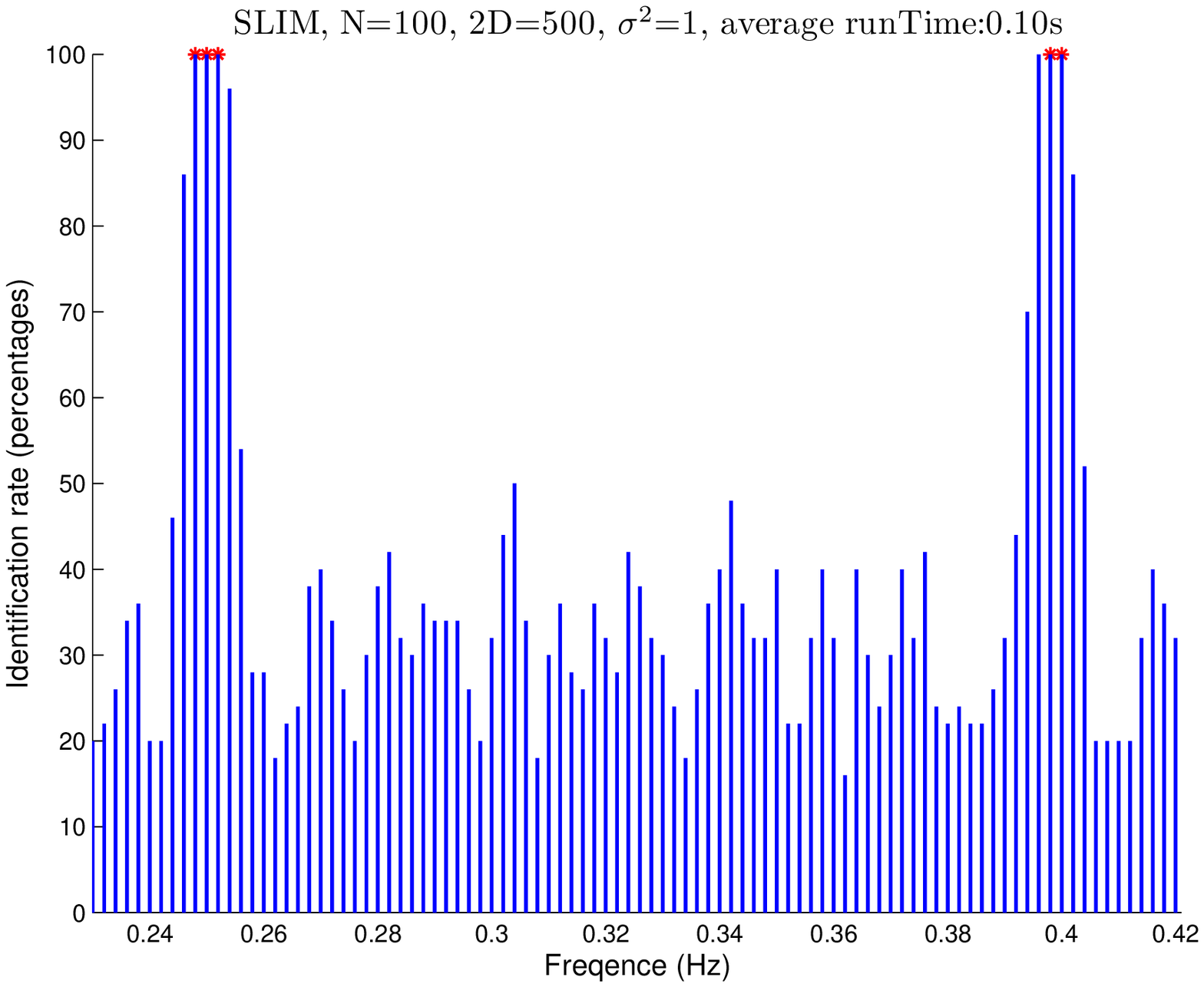}
    \caption{SLIM}
                    \label{subfig:SLIM_iden1}
\end{subfigure}

\begin{subfigure}[b]{0.4\textwidth}
    \centering\includegraphics[width=\textwidth]{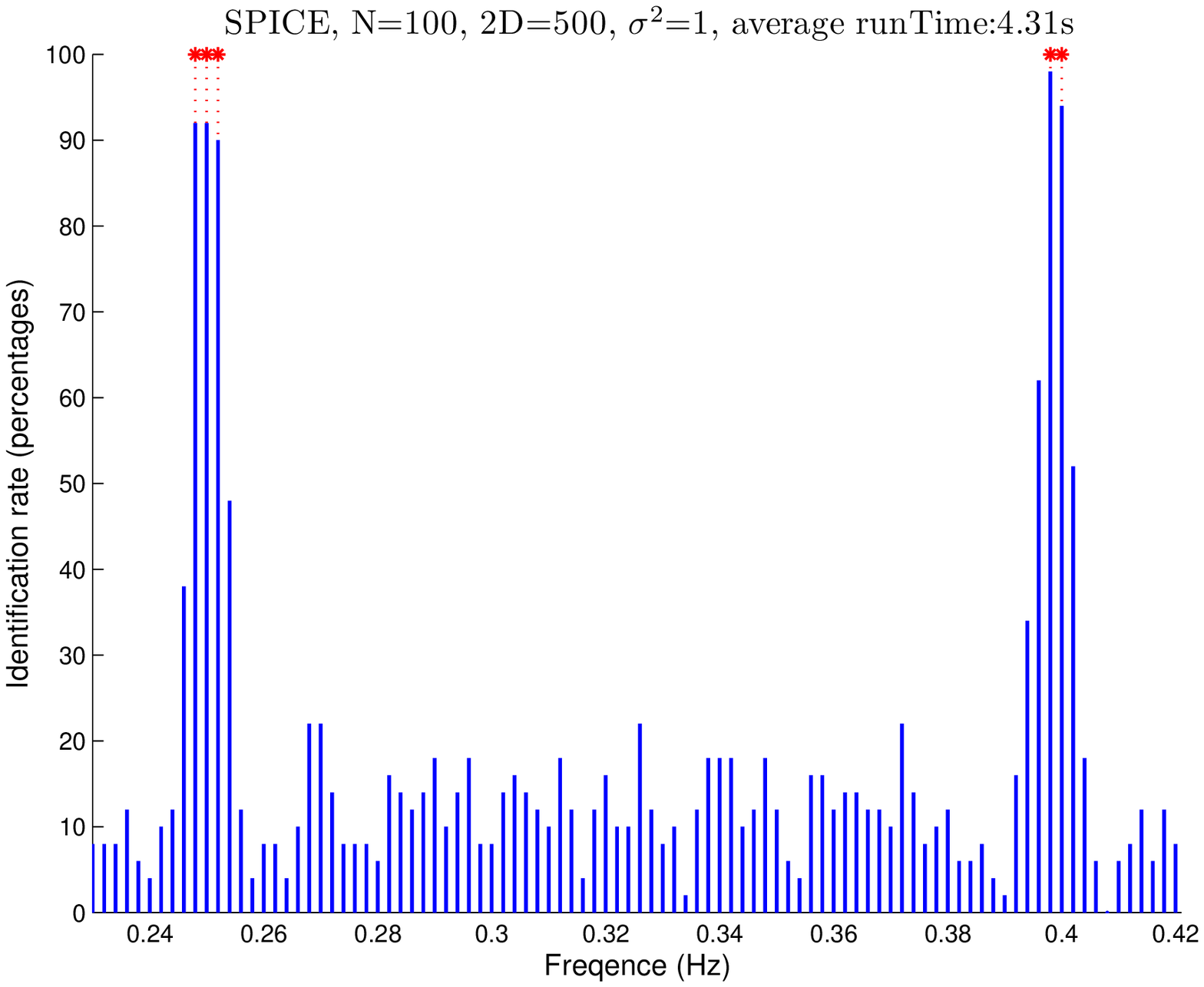}
    \caption{SPICE}
            \label{subfig:SPICE_iden1}
\end{subfigure}\hfill
\begin{subfigure}[b]{0.4\textwidth}
    \centering\includegraphics[width=\textwidth]{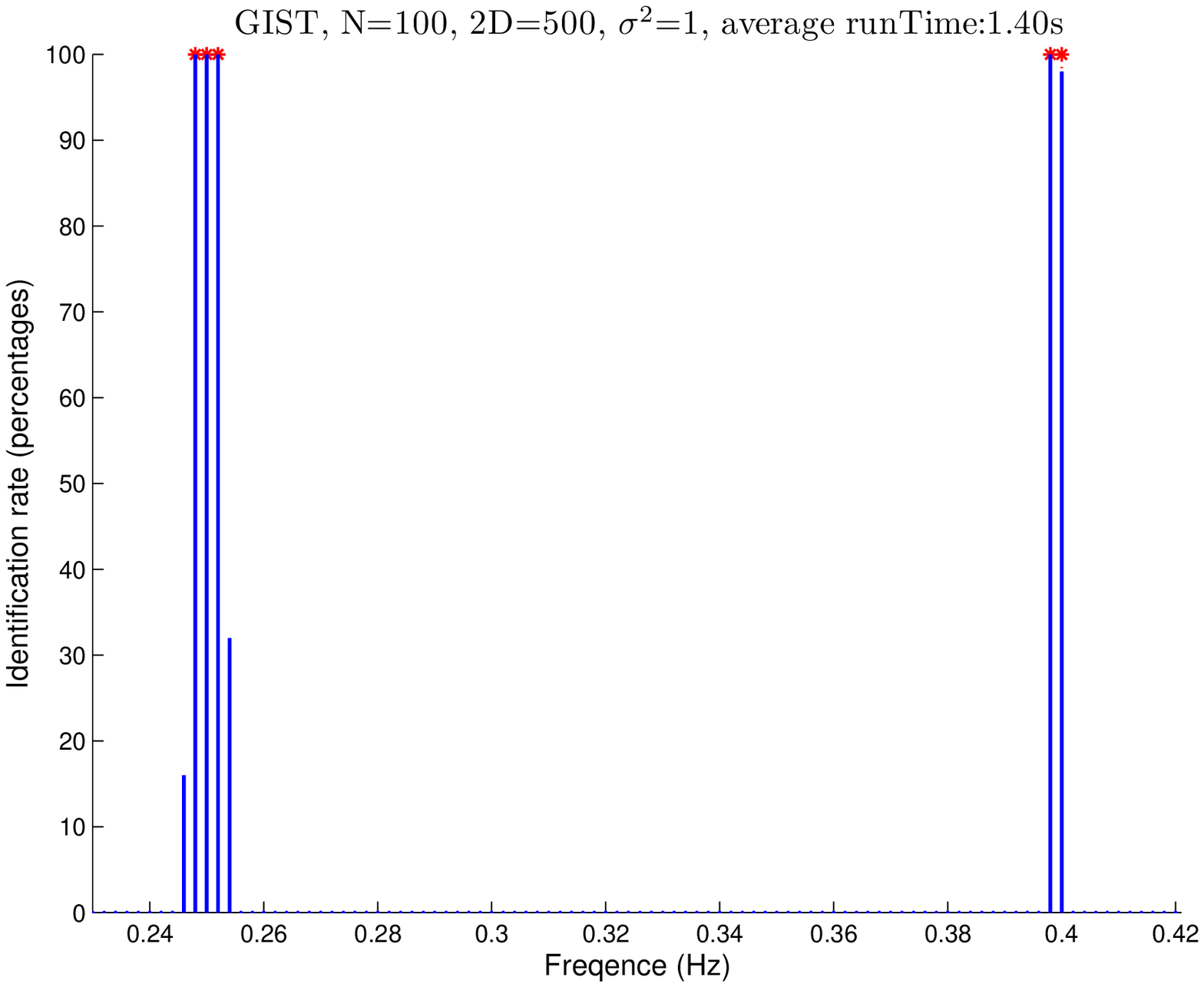}
    \caption{GIST}
                    \label{subfig:GIST_iden1}
\end{subfigure}

\caption{{
Frequency identification rates with $\sigma^2=1$ in 50 simulation runs, using BP, LZA-F, IAA, SLIM, SPICE, and GIST.}}
\label{fig:iden_results_sigmasq1}
\end{figure*}

\begin{figure*}[h!]
\centering

\begin{subfigure}[b]{0.4\textwidth}
    \centering\includegraphics[width=\textwidth]{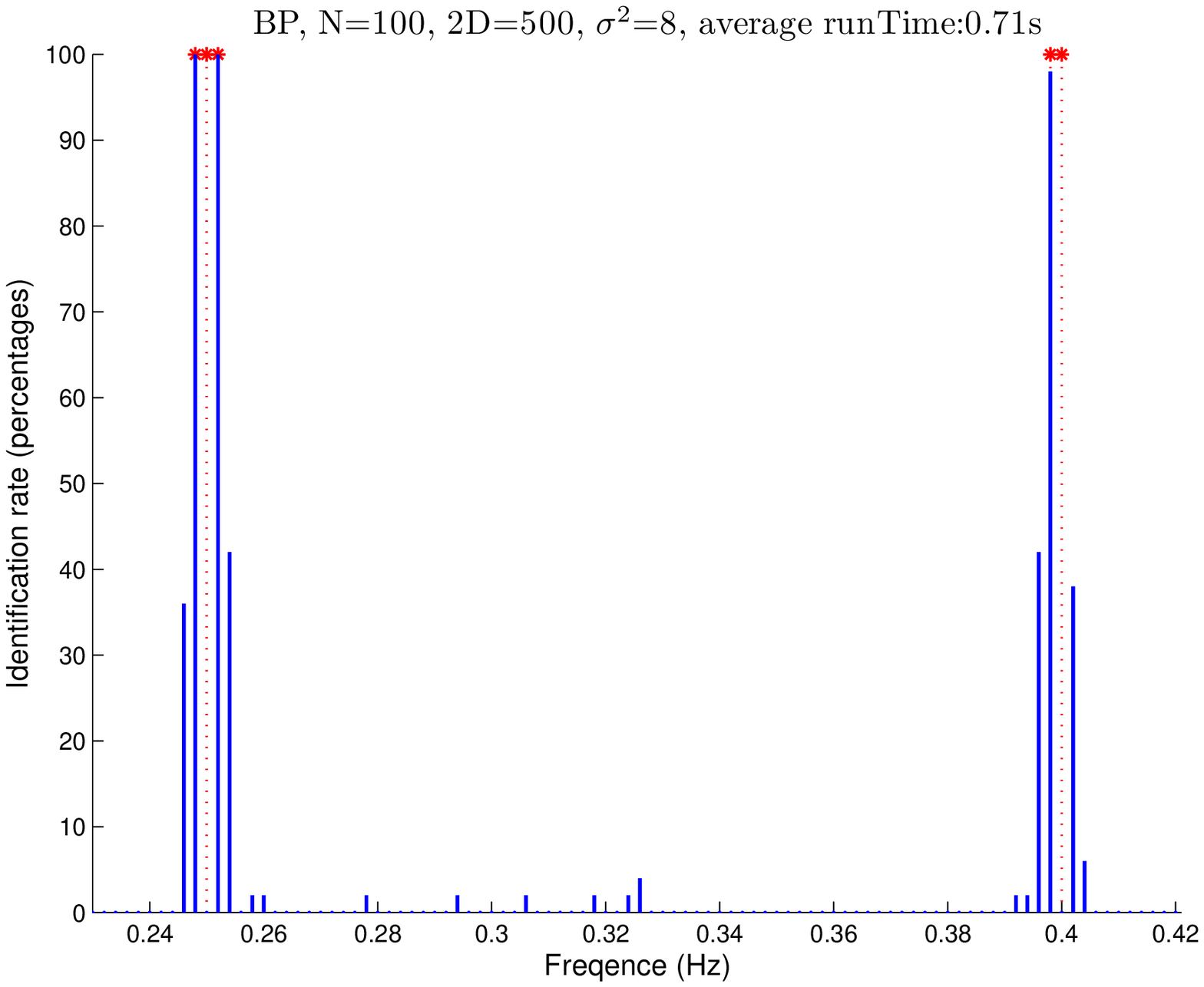}
    \caption{BP}
            \label{subfig:BP_iden8}
\end{subfigure}\hfill
\begin{subfigure}[b]{0.4\textwidth}
    \centering\includegraphics[width=\textwidth]{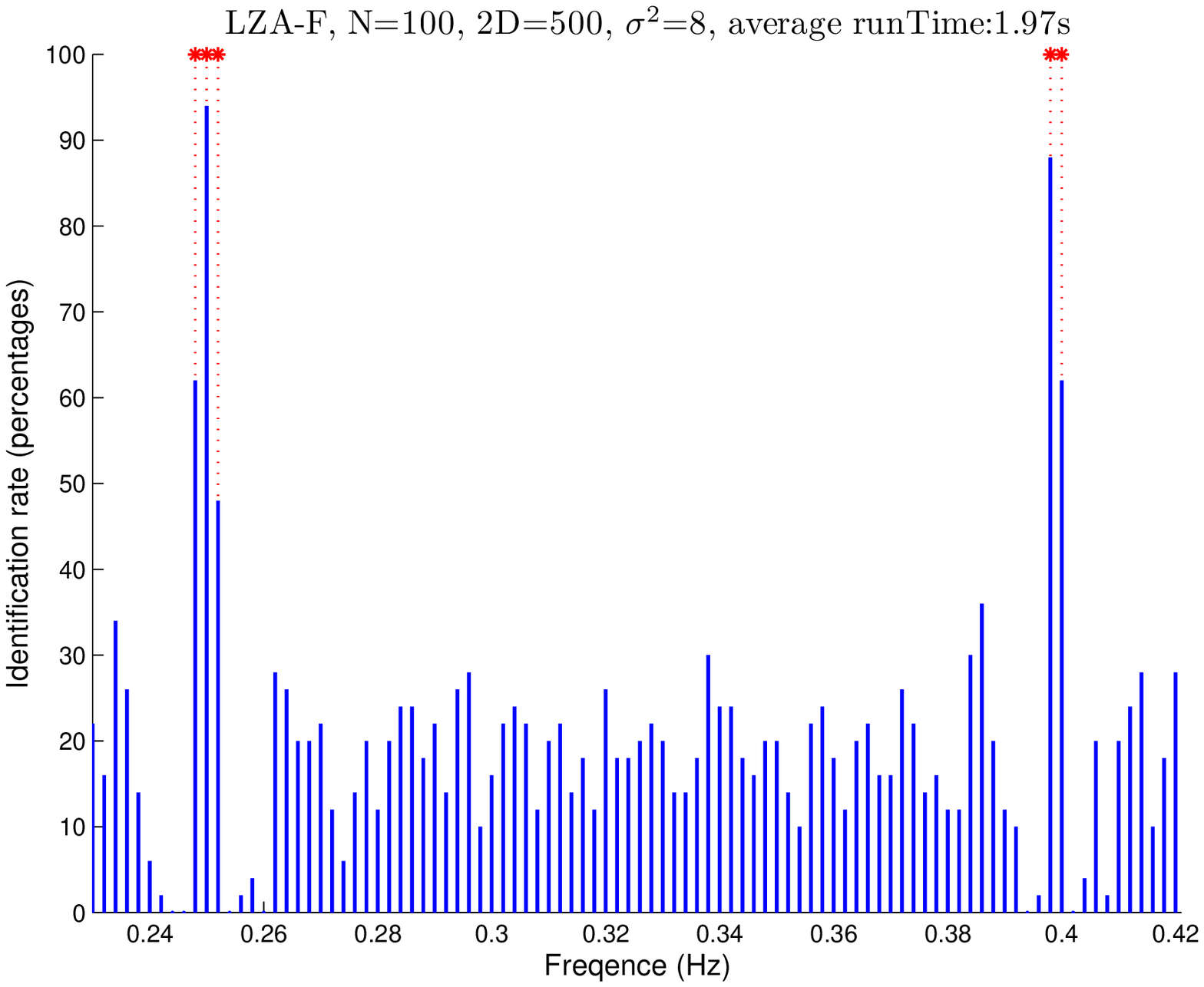}
    \caption{LZA-F}
                    \label{subfig:LZAF_iden8}
\end{subfigure}

\begin{subfigure}[b]{0.4\textwidth}
    \centering\includegraphics[width=\textwidth]{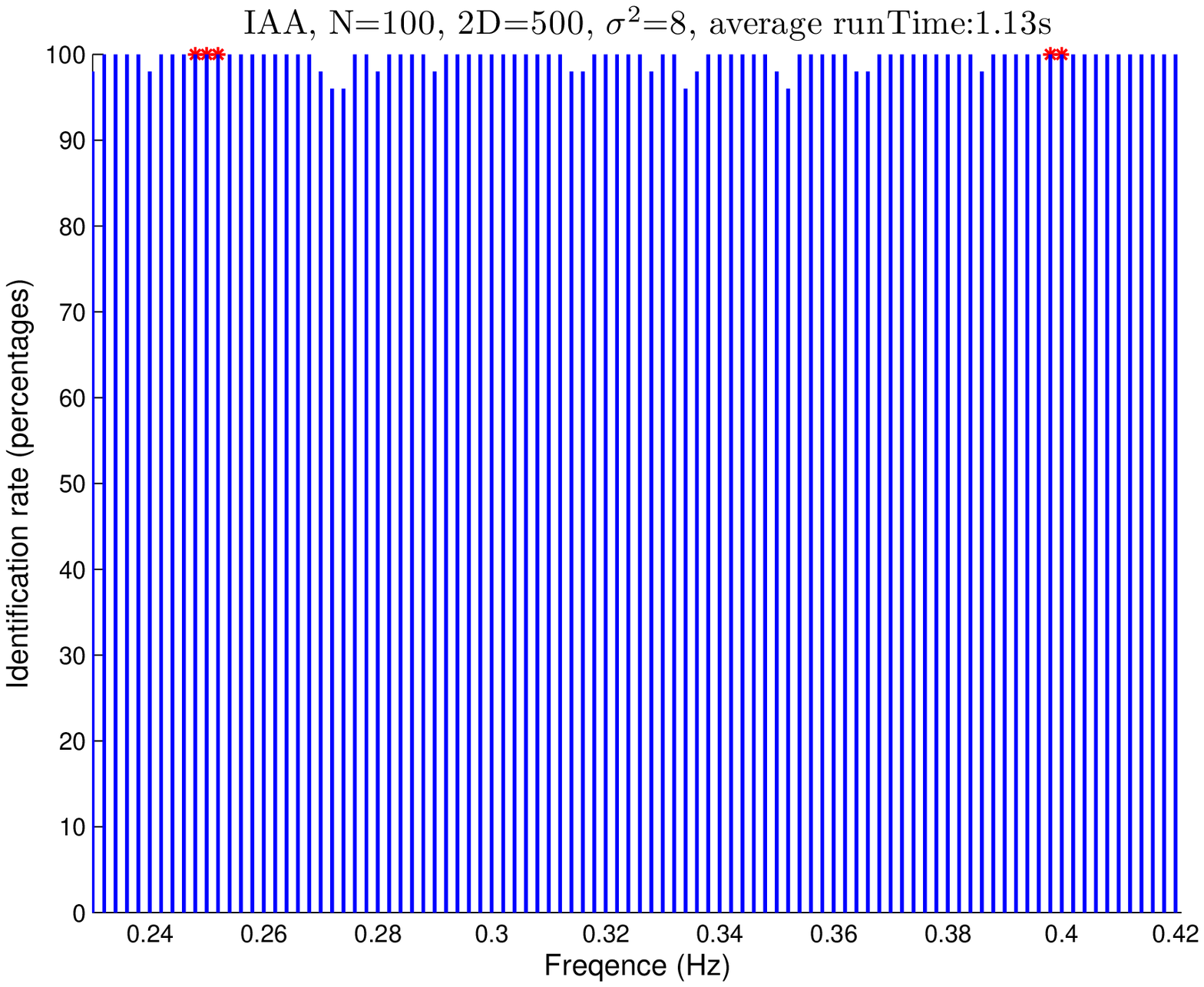}
    \caption{IAA}
            \label{subfig:IAA_iden8}
\end{subfigure}\hfill
\begin{subfigure}[b]{0.4\textwidth}
    \centering\includegraphics[width=\textwidth]{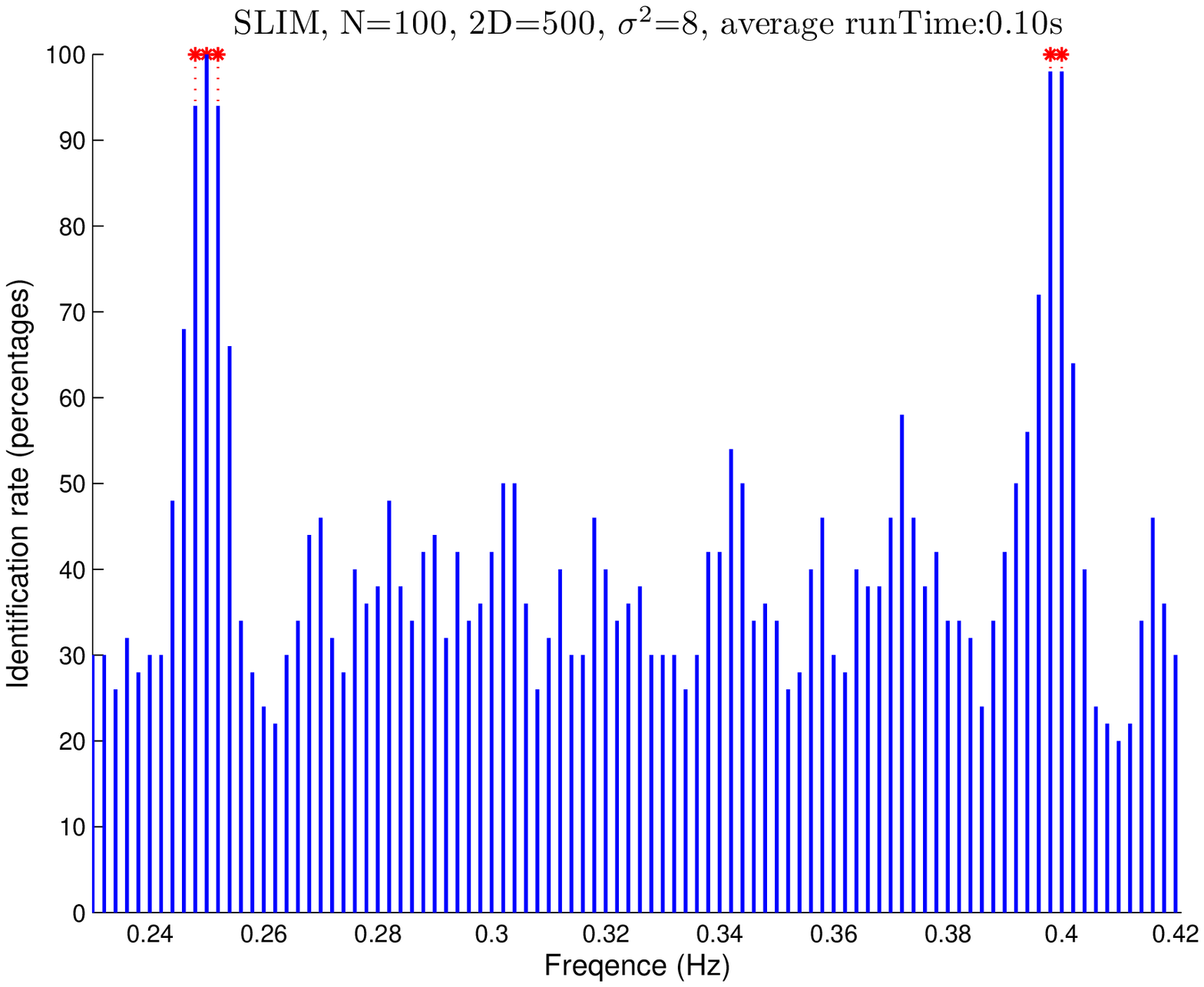}
    \caption{SLIM}
                    \label{subfig:SLIM_iden8}
\end{subfigure}

\begin{subfigure}[b]{0.4\textwidth}
    \centering\includegraphics[width=\textwidth]{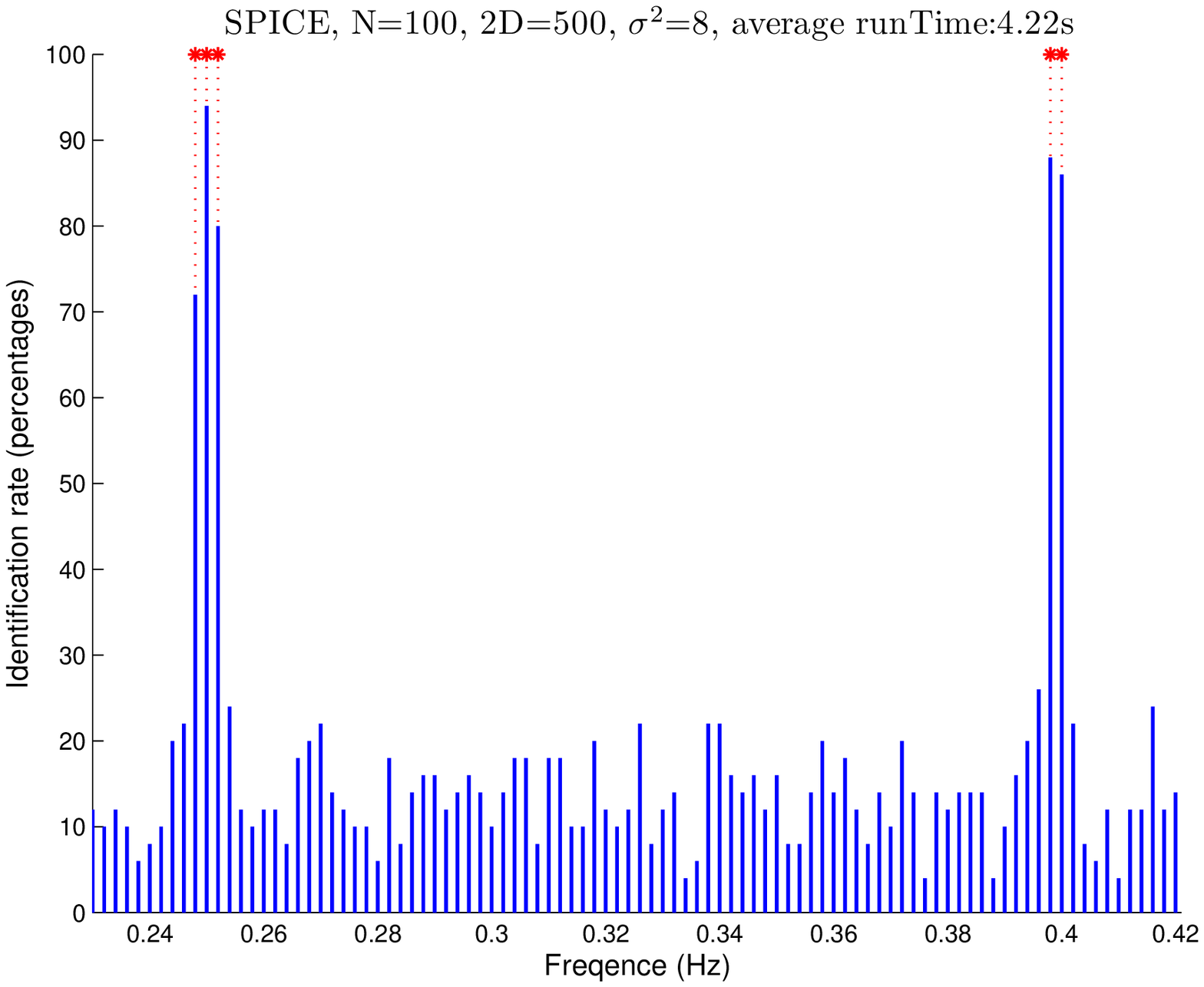}
    \caption{SPICE}
            \label{subfig:SPICE_iden8}
\end{subfigure}\hfill
\begin{subfigure}[b]{0.4\textwidth}
    \centering\includegraphics[width=\textwidth]{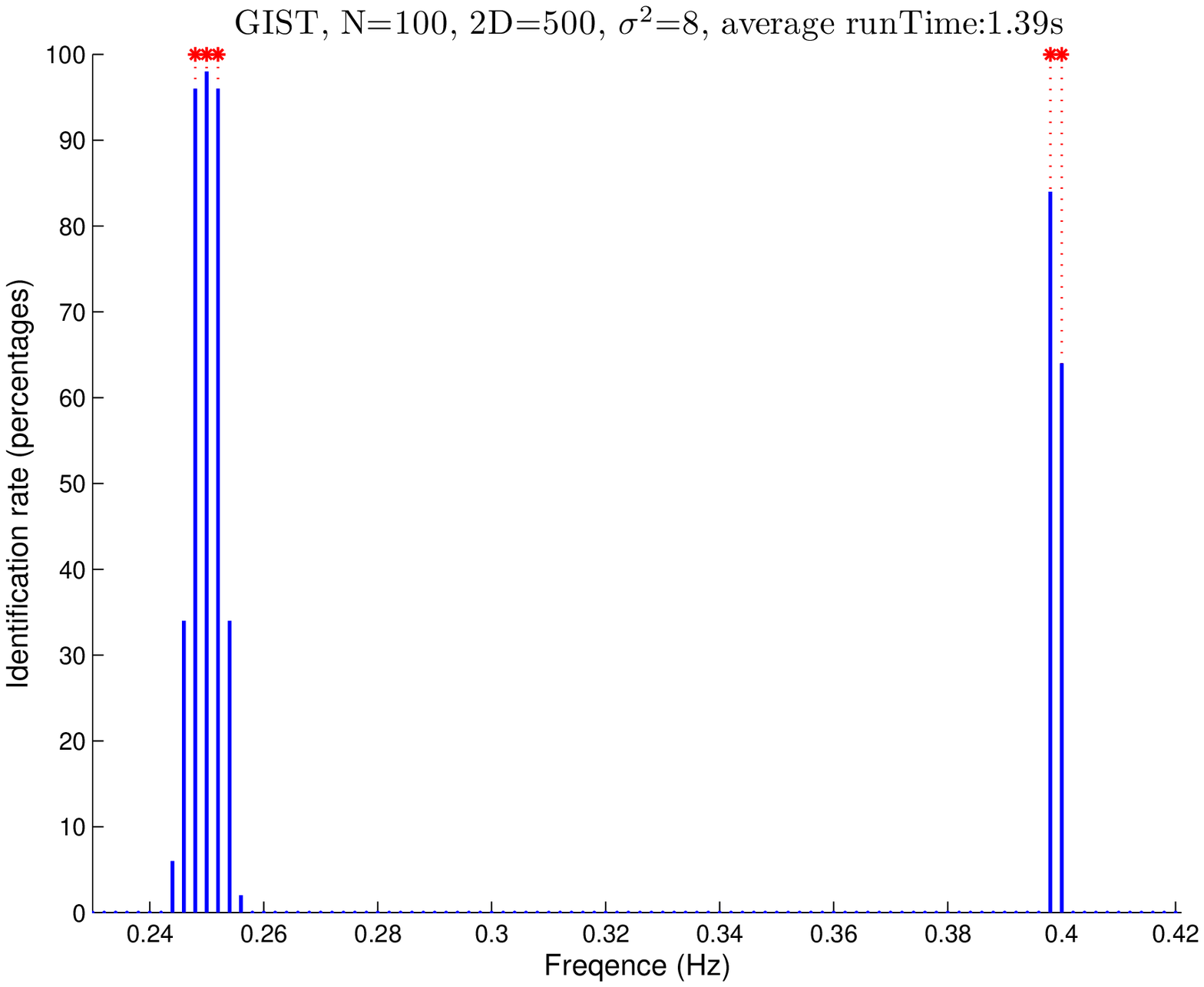}
    \caption{GIST}
                    \label{subfig:GIST_iden8}
\end{subfigure}

\caption{{
Frequency identification rates with $\sigma^2=8$ in 50 simulation runs, using BP, LZA-F, IAA, SLIM, SPICE, and GIST.}}
\label{fig:iden_results_sigmasq8}
\end{figure*}

To compare with the advocated \textbf{group hard-ridge GIST} (or GIST for short), we implemented   BP \cite{chen1998application}, IAA-APES (or IAA for short) \cite{iaa-apes}, SPICE \cite{spice},  LZA-F \cite{lza-f}, CG-SLIM (or SLIM for short) \cite{SLIM-cg}.   
To make a fair and realistic comparison, we used a common stopping criterion:  the number of iterations reaches $200$ or the  change  in $\bsbb$ is less than  1e-4.
In GIST, we set $\vartheta=0.25$ to give the cardinality bound in screening, and used SCV-BIC for parameter tuning.
The algorithmic  parameters in the other methods took default values suggested in  the literature. (For example, the $q$ parameter in SLIM  is chosen to be $1$, as recommended and used in the numerical examples of \cite{SLIM-cg}.)
Figs.~\ref{fig:iden_results_sigmasq1} and \ref{fig:iden_results_sigmasq8} show the frequency identification rates  in 50 simulation runs  for each of the methods under  $\sigma^2=1$ and  $\sigma^2=8$, respectively. That is, given each algorithm, we plotted the percentage of identifying $f_k$ or $\hat\bsbb_{f_k}\neq 0$ in all runs, for every $f_k$ in the dictionary. The blue solid lines show such identification rates, while the  red dotted lines (with star marks at $100\%$) label the true frequencies.
The plot for $\sigma^2=4$ is similar to Fig.~\ref{fig:iden_results_sigmasq8}; we do not show it here due to the page limit.
 We also included  the running time (averaged over 50 runs) in Table \ref{tab:timecomp} to reflect the computational cost.

%
%
%
%
%
%
%

\begin{table}[htb]
\renewcommand{\arraystretch}{1.5}
\centering
\caption{Average
runtime  in seconds of different algorithms, with varying values of $\sigma^2$ at $1$, $4$, and $8$.  }
  \label{tab:timecomp}
\begin{tabular}{r c c c}
&
\multicolumn{1}{c}{
~$\sigma^2=1$} &
\multicolumn{1}{c}{
~$\sigma^2=4$} &
\multicolumn{1}{c}{
~$\sigma^2=8$}\\
\hline

BP              & 0.80     & 0.72      & 0.71 \\
LZA-F          & 1.64    & 1.88     & 1.97     \\
\hdashline[1pt/2pt]
IAA           & 1.18   & 1.09  & 1.13 \\
SLIM           & 3.77    & 3.71   & 3.70  \\
SLIM with \emph{FFT} & 0.10    & 0.10   & 0.10  \\
\hdashline[1pt/2pt]
SPICE           & 4.31    & 4.23  & 4.22  \\
GIST            & 1.40     & 1.40    & 1.39  \\
\hline
\end{tabular}
\end{table}

IAA, SPICE,  LZA-F, and SLIM are \emph{not} capable of producing {inherently sparse} estimates. One must make a somewhat ad-hoc choice of the cutoff value $\tau$ to discern   the present frequencies.  We set $\tau=1e-2$ in performing such  post-truncation. It behaved better than $\tau=1e-3$ or $\tau=1e-4$ in experimentation  (which gave similar yet worse detection performance).

BP, though super fast, missed the frequency components at $0.25$ and $0.4$ all the time. An improvement  is offered by the CG-SLIM which makes use of the group $l_1$ regularization.
In \cite{SLIM-cg}, CG-SLIM is recommended to run for only 20 iteration steps (whereas the simulated signals there had  very mild noise contamination, with  $\sigma^2=0.001$).  Here, we increased the maximum number of iterations to 200 for better identification, without sacrificing much efficiency. Otherwise CG-SLIM gave much poorer spectrum recovery in experiments.

SLIM is free of parameter tuning, because from a  Bayesian perspective SLIM estimates the noise variance $\sigma^2$ in addition to the coefficient vector $\bsbb$. Unfortunately,  we found all the variance estimates from SLIM were  severely  biased downward---for example, for $\sigma^2=8$, the mean $\hat \sigma^2$ in 50 runs is about $3e-5$.  This is perhaps the reason why SLIM failed in super-resolution recovery: with such a small $\sigma^2$ estimate, the threshold level tends to be very low, and thus SLIM always overselects. 
Seen from the  figures, SLIM results in many spurious frequencies, some arising in more than 60 percent of the datasets.
IAA is even worse and does not seem to have the ability to super-resolve.

It is observed that LZA-F may seriously mask the true components. In addition, with moderate/large noise contamination, we found that LZA-F may be unstable and produce huge errors. Because the design of LZA-F is to approximate the $l_0$ regularization,  we  substituted the hard-thresholding for  $\Theta$ in GIST, which solves the exact $l_0$-penalized  problem. However, the high miss rates of the $l_0$-type regularization are still commonly seen, and the resulting models are often over-sparse. 
To give an explanation of this under-selection, notice that the $l_0$ regularization either kills or keeps, thereby offering {no} shrinkage at all for nonzero coefficients.
To attain the appropriate extent of shrinkage especially when the noise is not too small, it has  to kill more predictors than necessary.
As a conclusion,  inappropriate nonconvex penalties may seriously mask true signal components.

In our experiments, SPICE performs well.
GIST is much better and shows more concentrated signal power at the true frequencies.
It produces very few spurious frequencies, and in terms of computation, it is much more efficient than SPICE (Table \ref{tab:timecomp}).
GIST adapts to SNR and is both stable and scalable.

Finally,  a recent proposal of  using the  \emph{FFT} for matrix-vector multiplication  \cite{SLIM-cg} was shown to be very effective:  for SLIM, the average running time dropped from about 3.7 seconds to 0.1 seconds. The computational trick can be  applied to all of the methods discussed here.
However, it   restricts to  {uniformly} sampled data with  {Fourier} dictionaries. We did not use the FFT implementation for the other methods. (GIST algorithms and analyses are general and do not have such restrictions,  see Section \ref{sec:Model} and Section \ref{sec:Conclusions}.)

\subsubsection{Probabilistic spectral screening}
\label{subsec:ExptScreening}

\begin{figure*}[htp]
  \centering
  \includegraphics[width=0.6\textwidth]{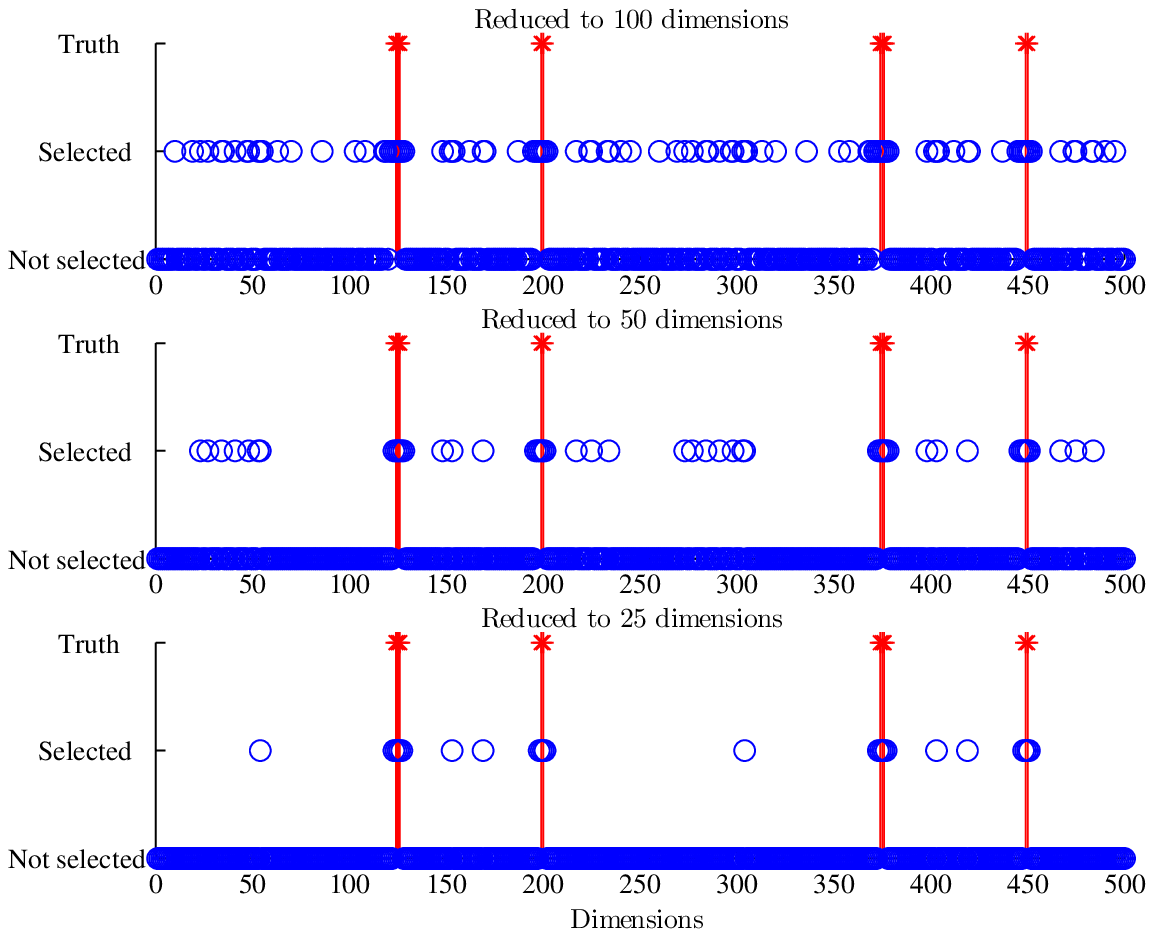}
  \caption[]{Locations of the remaining frequency atoms after GIST screening with $\vartheta N=100, 50, 25$. The true frequencies are indicated by red lines and stars.   
  }
  \label{fig:screened_500}
\vspace{.4in}
\begin{subfigure}[b]{0.5\textwidth}
    \centering\includegraphics[width=\textwidth]{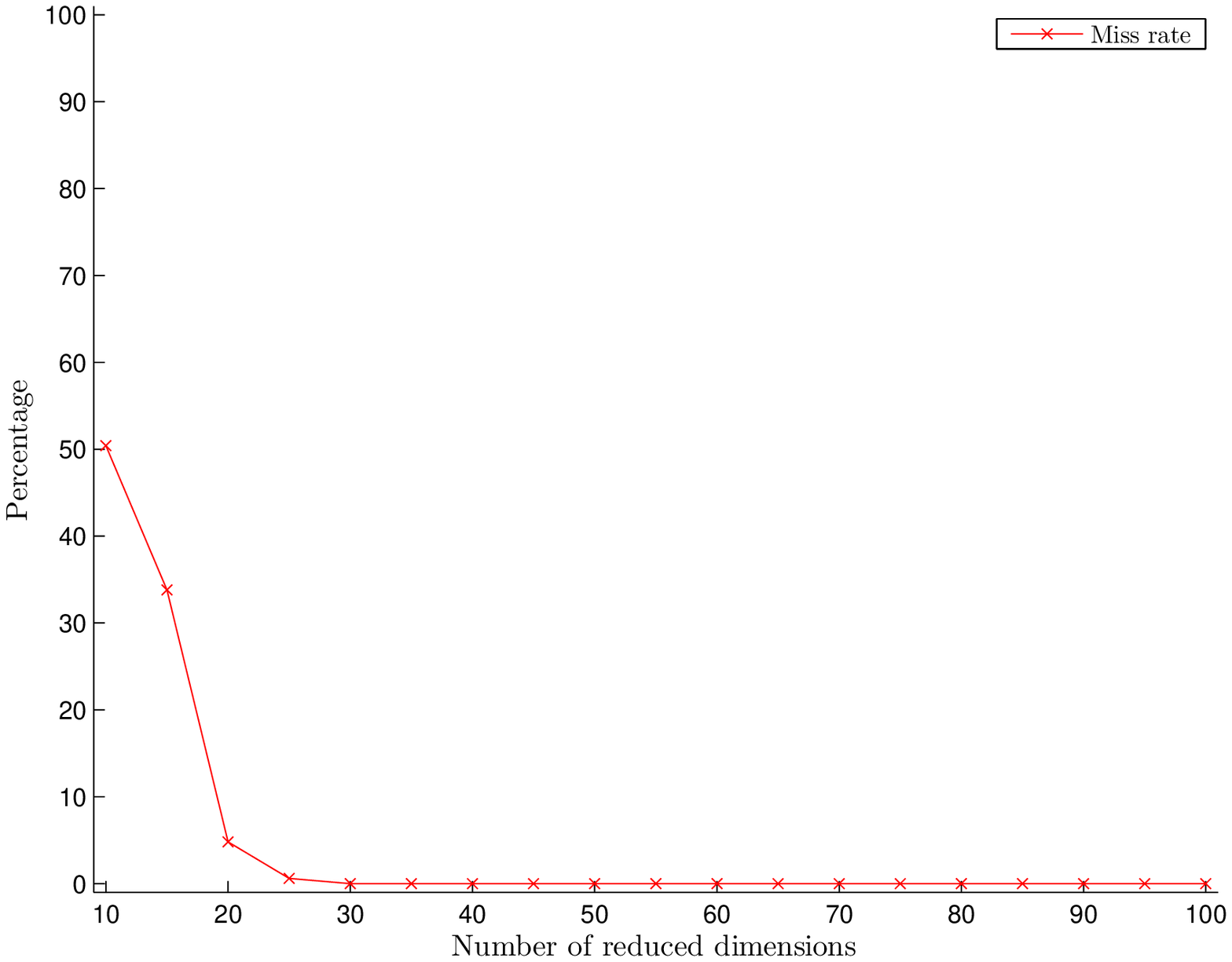}
             \label{subfig:scrtime}
\end{subfigure}\hfill
\begin{subfigure}[b]{0.5\textwidth}
    \centering\includegraphics[width=\textwidth]{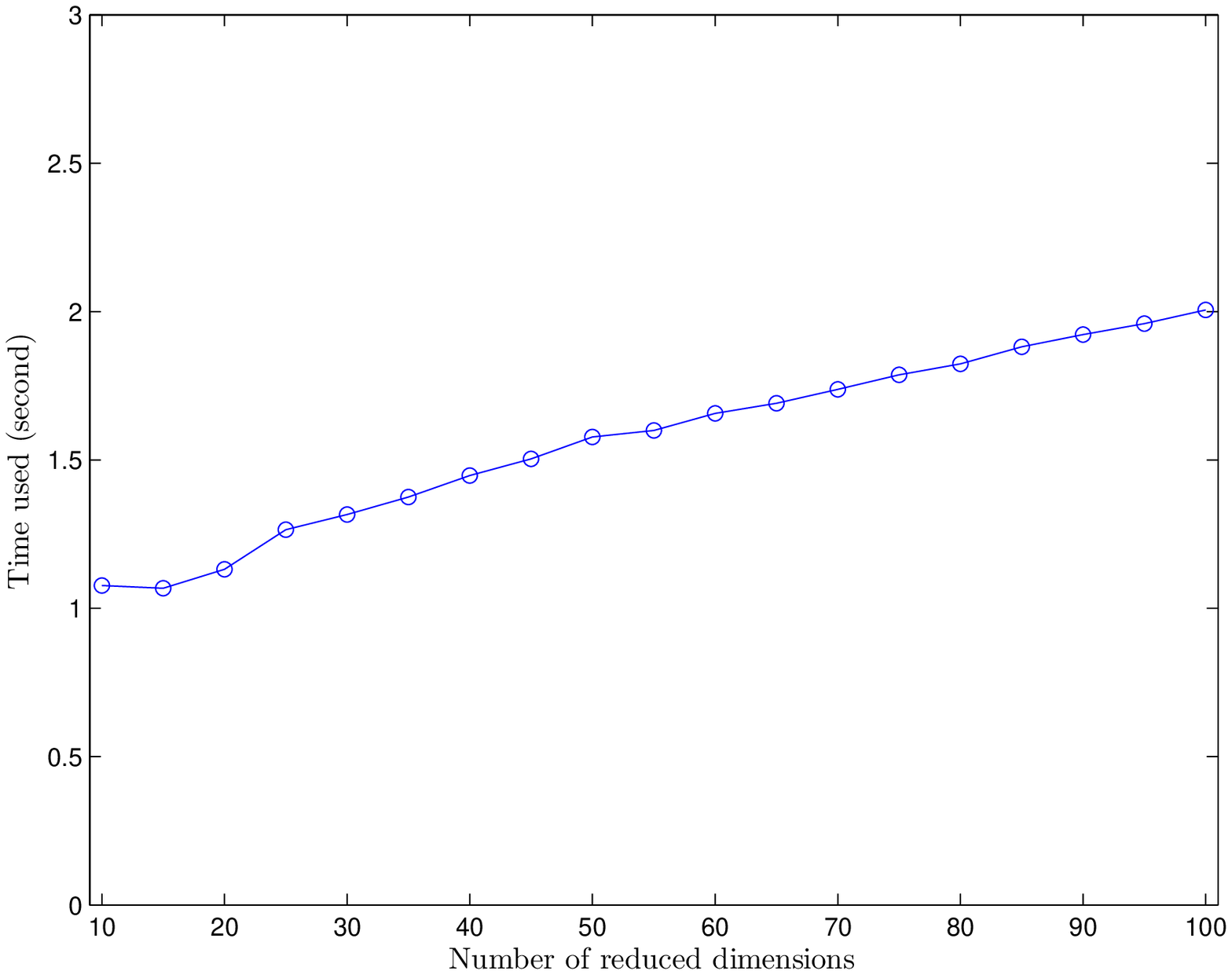}
            \label{subfig:scrperf}
\end{subfigure}
\caption[]{Performance of GIST screening on a hard problem with 10 present frequency components and large noise variance  $\sigma^2=10$. The left panel shows the miss rates, while the right panel shows the total computational time (including the  GIST fitting time thereafter); both $x$-axes represent $\vartheta N$, the dimensions to be kept after screening. }\label{fig:screening}
\end{figure*}

We   examine  the performance of the GIST-screening Algorithm \ref{alg:GIST-screening} in this experiment. 
The candidate ratio $\vartheta$ determines the dimensions ($\vartheta N$) of the reduced predictor space. Therefore, the lower the value of $\vartheta$, the more efficient the computation, but also the higher the risk of mistakenly removing some true components.
Our screening technique turns out to be pretty successful:  even if we choose $\vartheta N$ to be as small as $25$ (which can be even lower), 
it never misses any true frequency component. 
Fig. \ref{fig:screened_500} shows the frequency location of $100$, $50$, and $25$ remaining  atoms, respectively, in GIST screening. The selected frequencies are {non-uniform}, and the density near the true spectra is much higher.

Next, we make a much more challenging problem by modifying the signal to have  10 present frequency components at $0.24, 0.242,  \ldots,  0.282$ and large noise variance $\sigma^2=10$.
Fig.~\ref{fig:screening} shows both the detection miss rates and the computation time, averaged over 50 runs. 
The {miss} rate is  the mean of $| \{ i: \beta_i^* \neq 0,\hat{\beta}_i=0\} | / | \{ i: \beta_i^* \neq 0\} | $ in all simulations, where $|\cdot|$ is the cardinality of a set.
The plotted time is the total running time of both GIST screening and model fitting and selection.
The empirical experience is  that GIST-screening is safe when $\vartheta N$ is roughly 3 times greater than the number of truly relevant atoms. 
It reduces the computation complexity significantly with little performance lost. 

\subsubsection{Misspecified resolution level}
\label{subsec:ExptNongrid}
In  super-resolution spectral selection, the frequency resolution level $\delta$  used in dictionary construction is customized by users. This requires the knowledge of a lower bound on frequency spacing.  We are particularly interested in the performance of GIST when  $\delta$ is misspecified in reference to the truth.

In this experiment, we set the signal frequencies at  $0.2476$, $0.2503$, $0.2528$,  $0.3976$, $0.4008$, with   amplitudes $A_k$ and phases $\phi_k$  unchanged.
Clearly, the ideal frequency resolution to resolve this signal should be no more than $0.0001$ Hz.

We chose $\delta=0.002$,  20 times as large as the required resolution.
The results are nearly identical to Figs. \eqref{subfig:GIST_iden1}, \eqref{subfig:GIST_iden8} (not shown due to space limitation).
The crude resolution specification makes GIST unable to recover the true frequencies.
On the other hand, the most frequently identified frequencies  are $0.248$, $0.25$, $0.252$, $0.398$, $0.4$, and a comparison shows that this is the best approximation in the given frequency grid. (For example, $0.398$Hz  is the closest frequency in the grid $\{0, 0.02, \cdots, 0.396, 0.398, \cdots, 0.5\}$ to $0.3976$Hz.)
This phenomenon is also seen in many other experiments:
GIST gives the best possible identification to approximate the true frequencies,  with the quantization error determined by the resolution level.

\section{Conclusions}
\label{sec:Conclusions}

We have presented a sparsity-based GIST framework to tackle the   super-resolution challenge in spectral estimation. It is able  to handle nonconvex penalties  and  take  the pairing structure of sine and cosine atoms into account in regularizing the model. The $l_0+l_2$ type hard-ridge penalty was shown to be able to dramatically improve   the popular  convex $l_1$ penalty as well as  the nonconvex $l_0$ penalty.
Its variant, the iterative probabilistic spectrum screening,  can be used for supervised dimension reduction and fast computation. In parameter tuning, the SCV   criterion overcomes the training inconsistency issue of the plain CV and is much more computationally efficient.
GIST can be applied to unevenly sampled signals (in which case the sampling time sequence $\{t_n\}_{1\leq n \leq N}$ is not  uniform) with guaranteed convergence (cf. Theorem \ref{th:conv} and Theorem \ref{th:scr}).

It is worth mentioning that in  our algorithm design and theoretical analyses, the only use of the Fourier frequency dictionary was to extract the atom grouping manner.  Our methodology carries over to any type of dictionary as arising in signal processing, wavelets, and statistics.
For example, although we focused on real-valued signals in the paper,  for complex-valued signals, say, $\bsby = [y(t_n)] \in \mathbb C^{N\times 1}$  observed at $t_n$ ($1\leq n \leq N$) and  the candidate frequency grid  given by $f_k$ ($1\leq k \leq D$), a complex dictionary $\bsbX$ can be constructed as $[\exp(i 2\pi  f_k t_n)] \in \mathbb C^{N\times D}$ in place of \eqref{eqn:X0}, with  $\bsbb=[\beta_1, \cdots, \beta_D]^T\in \mathbb C^D$.
 The  group penalized model  then minimizes 
$
 {1\over2}\|\bsby - \alpha-\bsbX \bsbb \|_2^2 + \sum_{k=1}^D P\left(\|\beta_k\|_2; \lambda\right)
$
where $\| \beta_k\|_2 = \sqrt{\mbox{Re}(\beta_k)^2 + \mbox{Im}(\beta_k)^2}$, simply the complex norm of $\beta_k$.
It is straightforward to extend all our algorithms and analyses to this problem. On the other hand, for real-valued signals, the formulation using the sine-cosine predictor matrix in \eqref{eqn:X0}  does not involve any imaginary/complex number processing in implementation.

Some future research topics include the extension of GIST to non-Gaussian and/or multivariate signals.


\appendices
\section{Proof of Theorem \ref{th:conv}}
\label{app:proofconv}
We show the result for the group form only. The proof for the non-group form is similar and simpler.
The following \textbf{continuity assumption} is made throughout the  proof:

\emph{Assumption $\mathfrak{A}$}: \emph{$\vec \Theta$  is continuous at any point in the closure of $\{\bsbxi^{(j)}\}$. }
\\
For continuous thresholding rules such as soft-thresholding, this regularity condition  always holds. Practically used thresholding rules (such as hard-thresholding) have  few discontinuity points and such discontinuities rarely occur in any real application.
For $\omega=1$, see \cite{SheGLMTISP} for the proof details. In the following, we assume $0< \omega < 1$.

Note that $G$ is quadratic and convex in $\bsbb$, $\bsbxi$, and $\bsbzeta$, but possibly nonconvex and nonsmooth in $\bsbg$.

\begin{lemma}
\label{uniqsol-gen-grp}
Given an arbitrary thresholding rule $\Theta$,
let $P$ be any function satisfying
$$P(\theta;\lambda)-P(0;\lambda)=P_{\Theta}(\theta; \lambda) + q(\theta; \lambda)$$
where $P_{\Theta}(\theta; \lambda)\triangleq \int_0^{|\theta|} (\sup\{s:\Theta(s;\lambda)\leq u\} - u) \rd u$, $q(\theta; \lambda)$ is nonnegative and $q(\Theta(t;\lambda))=0$ for all $t$.
Then, the minimization problem
\begin{align*}
\min_{\bsbb }  \frac{1}{2}\|\bsby-\bsbb\|_2^2 + P(\|\bsbb\|_2;\lambda)
\end{align*}
%
has a unique optimal solution given by $\hat\bsbb=\vec\Theta(\bsby;\lambda)$ for every $\bsby$ provided that $\Theta(\cdot;\lambda)$ is continuous at $\|\bsby\|_2$.
\end{lemma}

See  \cite{SheGLMTISP} for its proof.

Given $\bsbb$ and $\bsbxi$, the problem of minimizing $G$ over $(\bsbg, \bsbzeta)$ can be simplified to (detail omitted)
$$
\min_{\bsbg} \frac{1}{2} \| \bsbg - \omega ( \bsbI - \bsbSig) \bsbb - \omega \bsbX^T \bsby - (1-\omega) \bsbxi \|_2^2 + P(\bsbg;\lambda),
$$
 and
\begin{align*}
\min_{\bsbzeta}   \frac{1}{2} \frac{1-\omega}{\omega} [ \bsbzeta - \omega(\bsbI - \bsbSig) \bsbb - \omega \bsbX^T \bsby - (1-\omega) \bsbxi]^T ( \bsbI - \bsbSig)^{-1}
[ \bsbzeta - \omega(\bsbI - \bsbSig) \bsbb - \omega \bsbX^T \bsby - (1-\omega) \bsbxi].
\end{align*}
Based on Lemma \ref{uniqsol-gen-grp}, the optimal solutions are
\begin{align*}
\begin{cases}
\bsbg_{opt}&=\vec\Theta(\omega ( \bsbI - \bsbSig) \bsbb + \omega \bsbX^T \bsby + (1-\omega) \bsbxi; \lambda)\\
\bsbzeta_{opt}&=\omega ( \bsbI - \bsbSig) \bsbb + \omega \bsbX^T \bsby + (1-\omega) \bsbxi.
\end{cases}
\end{align*}
 Therefore, we obtain
\begin{align}
G(\bsbb^{(j+1)}, \bsbxi^{(j+1)}, \bsbb^{(j)}, \bsbxi^{(j)}; \lambda) \leq G(\bsbb^{(j)}, \bsbxi^{(j)}, \bsbb^{(j)}, \bsbxi^{(j)}; \lambda) - \frac{1-\omega}{2\omega} (\bsbxi^{(j+1)} - \bsbxi^{(j)})^T ( \bsbI - \bsbSig)^{-1} (\bsbxi^{(j+1)} - \bsbxi^{(j)}).\label{dec1}
\end{align}

On the other hand, given $\bsbg$ and $\bsbzeta$, $G$ can be expressed as a  quadratic form in $\bsbb$ that is positive definite.  The same fact holds for  $\bsbxi$.  It can be computed that
\begin{align*}
\begin{cases}
\nabla G_{\bsbb}  = \omega ( \bsbI - \bsbSig)  (\bsbb - \bsbg) + (1-\omega) (\bsbxi - \bsbzeta)\\
\nabla G_{\bsbxi}=\frac{1-\omega}{\omega} ( \bsbI - \bsbSig)^{-1} [ \omega ( \bsbI - \bsbSig) (\bsbb - \bsbg) + (1-\omega)  (\bsbxi - \bsbzeta)],
\end{cases}
\end{align*}
from which it follows that $G$ can be written as $\frac{1}{2} [\omega ( \bsbI - \bsbSig)  (\bsbb - \bsbg) + (1-\omega) (\bsbxi - \bsbzeta)]^T \omega^{-1} (\bsbI -\bsbSig)^{-1}  [\omega ( \bsbI - \bsbSig)  (\bsbb - \bsbg) + (1-\omega) (\bsbxi - \bsbzeta)]$ in addition to the terms involving only $\bsbg$ and $\bsbzeta$.
Hence $\bsbb_{opt}=\bsbb$ and $\bsbxi_{opt}=\bsbzeta$ (though not unique)  achieve the minimum. We obtain
$G(\bsbb^{(j+1)}, \bsbxi^{(j+1)}, \bsbb^{(j+1)}, \bsbxi^{(j+1)}; \lambda)   \leq   G(\bsbb^{(j+1)}, \bsbxi^{(j+1)}, \bsbb^{(j)}, \bsbxi^{(j)}; \lambda) - \frac{1}{2\omega} [\omega ( \bsbI - \bsbSig)  (\bsbb^{(j)} - \bsbb^{(j+1)}) + (1-\omega) (\bsbxi^{(j)} - \bsbxi^{(j+1)})]^T ( \bsbI - \bsbSig)^{-1}
 [\omega ( \bsbI - \bsbSig)  (\bsbb^{(j)} - \bsbb^{(j+1)}) + (1-\omega) (\bsbxi^{(j)} - \bsbxi^{(j+1)})].
$
Combining this with \eqref{dec1}  yields \eqref{fundecrease}.

Assume a subsequence $\bsbb^{(j_l)}\rightarrow\bsbb^{\circ}$ as $l\rightarrow\infty$. Because
\begin{align*}
G(\bsbb^{(j_l)}, \bsbxi^{(j_l)}, \bsbb^{(j_l)},  \bsbxi^{(j_l)}) - G(\bsbb^{(j_l+1)}, \bsbxi^{(j_l+1)}, \bsbb^{(j_l+1)},  \bsbxi^{(j_l+1)})
\longrightarrow 0,
\end{align*}
 we have  $\bsbxi^{(j_l)} - \bsbxi^{(j_l+1)} \rightarrow 0$ and thus $(\bsbb^{(j_l)} - \bsbb^{(j_l+1)})\rightarrow 0$.
That is,
$(1-\omega)\bsbxi^{(j_l)}+\omega ( \bsbb^{(j_l)}+ \bsbX^T(\bsby-\bsbX\bsbb^{(j_l)}) ) - \bsbxi^{(j_l)}  \rightarrow 0$ and
$\vec\Theta (\bsbxi^{(j_l)};\lambda)-\bsbb^{(j_l)}\rightarrow 0$. From   $\bsbX^T(\bsby-\bsbX\bsbb^{(j_l)})- \bsbxi^{(j_l)}\rightarrow 0$ and the continuity assumption,   $\bsbb^{\circ}$ is a group $\Theta$-estimate satisfying \eqref{thetaeq-grp}, and $\lim_{j\rightarrow\infty} G(\bsbb^{(j_l)}, \bsbxi^{(j_l)}, \bsbb^{(j_l)}, \bsbxi^{(j_l)}) = F(\bsbb^{\circ})$.

\section{Proof of Theorem \ref{th:sel}}
\label{app:proofselprob}

Recall that $\mathcal F$ denotes the frequency set covered by the dictionary $\bsbX$ and we assume all column norms of $\bsbX$ are  $\sqrt N$ (or
the diagonal entries of $\bsbSig=\bsbX^T\bsbX$ are equal to  $N$).

Applying Theorem \ref{th:conv}, we can characterize any group $l_1$ estimate  $\hat\bsbb$ from Algorithm \ref{alg:TISP for PLM}   by
\begin{align}
\hat\bsbb = \vec\Theta(\hat\bsbb + \bsbX^T \bsby/\tau_0^2 - \bsbSig\hat\bsbb/\tau_0^2; \lambda) \label{thetaeq-scaled}
\end{align}
with $\Theta$ being the  soft-thresholding function.
Let    $\bsbs=\bsbs(\bsbb)$ denote a function of $\bsbb$ 
 satisfying
 \begin{align}
\|\bsbs_{{f}}\|_2 \leq  1, \forall f \in z(\bsbb), \ \bsbs_{{f}} =\bsbb_{f}/\|\bsbb_{f}\|_2, \forall f\in  nz(\bsbb), \label{s-def}
\end{align}
and $\bsbs(\bsbb_{f}):=[\bsbs(\bsbb)]_{f}$.
We have  $\|\bsbs(\bsbb_f)\|_2 \leq 1$,  $\forall f \in \mathcal{F}$.
 (In the group $l_1$ case, $\bsbs$ is a subgradient of $\sum_{f\in\mathcal{F}} \|\bsbb_{f}\|_2$.) Then \eqref{thetaeq-scaled} reduces to
$\hat\bsbb + \lambda \bsbs(\hat\bsbb, \lambda) =  \hat\bsbb + \bsbX^T \bsby/\tau_0^2 - \bsbSig \hat \bsbb/\tau_0^2$ or
\begin{align}
\bsbSig \hat\bsbb = \bsbX^T \bsby - \lambda \tau_0^2  \bsbs(\hat\bsbb),  \label{thetaeq-2}
\end{align}
for some $\bsbs$ satisfying \eqref{s-def}.


\begin{lemma}
\label{dawl:lemma1}
Assume $\bsbSig_{{nz^*}}$ is nonsingular.
Then \eqref{thetaeq-2} is equivalent to
\begin{align}
\begin{cases}
\bsbS_{{z^*}}\hat \bsbb_{{z^*}} =   \bsbX_{z^*}^{'T} \bsb{e} +
\lambda \tau_0^2\bsbSig_{{z^*}, {nz^*}} \bsbSig_{{nz^*}}^{-1} \bsbs(\hat\bsbb_{{nz^*}}) - \lambda \tau_0^2
\bsbs(\hat\bsbb_{{z^*}})  \\
\hat \bsbb_{{nz^*}} = \bsbb_{{nz^*}}^* + \bsbSig_{{nz^*}}^{-1} ( \bsbX_{{nz^*}}^{T} \bsbe - \lambda \tau_0^2 \bsbs(\hat\bsbb_{{nz^*}})
) -\bsbSig_{{nz^*}}^{-1} \bsbSig_{{z^*}, {nz^*}}^T \hat\bsbb_{{z^*}} \label{transfkkt}
\end{cases}
\end{align}
where  $\bsbS_{{z^*}}:=\bsbSig_{{z^*}} - \bsbSig_{{z^*}, {nz^*}} \bsbSig_{{nz^*}}^{-1} \bsbSig_{{nz^*}, {z^*}}$, and $\bsbX_{z^*}^{'T}:=\bsbX_{{z^*}}^{T} - \bsbSig_{{{z^*}}, {{nz^*}}} \bsbSig_{{nz^*}}^{-1} \bsbX_{{nz^*}}^T $.
\end{lemma}

The proof details are given  in   \cite{she2009thresholding}.

\begin{lemma}\label{B.2}
Suppose $\left[\begin{array}{c} z_1\\ z_2 \end{array}\right] \sim   N\left(\left[\begin{array}{c} 0\\ 0 \end{array}\right], \bsbV\right)$, where  $\bsbV$ is a correlation matrix. Then for any $M$, $P(z_1^2+z_2^2 >  M^2)\leq P(\xi>M^2/2)$ with $\xi\sim \chi^2(2)$.
\end{lemma}

From  $\| \bsbV\|_2 \leq \|\bsbV\|_F\leq 2$, $2\bsbI - \bsbV$ is positive semi-definite. Let $z_1'$, $z_2'$ be independent standard Gaussian random variables. We get $P(z_1^2+z_2^2 >  M^2)\leq P((z_1' \sqrt{2})^2 + (z_2' \sqrt{2})^2 > M^2)=P(\xi > M^2/2)$ from Anderson's inequality \cite{And55}.

\begin{lemma}\label{B.3}
Suppose $\xi\sim \chi^2(2)$. Then for any $M$, $P(\xi > 2M^2)\leq M^2 e^{-(M^2-1)}$.
\end{lemma}

See, e.g., \cite{cava02} for a proof of this  $\chi^2$ tail bound.

Let $\bsbX_{f}^{'T}=\bsbX_{{f}}^{T} - \bsbSig_{{{f}}, {{nz^*}}} \bsbSig_{{nz^*}}^{-1} \bsbX_{{nz^*}}^T $, $\forall f \in z^*$.
From Lemma \ref{dawl:lemma1}, we have $P_1 \geq P(A\cap V)$, with
$
A := \{\| \bsbX_{{f}}^{'T}
\bsbe + \lambda \tau_0^2 \bsbSig_{{f}, {nz^*}} \bsbSig_{{nz^*}}^{-1} \bsbs(\hat\bsbb_{nz^*}) \|_2 \leq \lambda \tau_0^2, \forall \bsbs \mbox{ satisfying \eqref{s-def}}, \forall f\in z^* \}
$,
$V:= \{ \|[\bsbSig_{{nz^*}}^{-1} \bsbX_{{nz^*}}^{T} \bsbe]_f \|_2 +  \lambda \tau_0^2 \| [\bsbSig_{{nz^*}}^{-1}
\bsbs(\hat\bsbb_{nz^*})]_f  \|_2 < \left\| \bsbb_{f}^* \right\|_2, \forall \bsbs \mbox{ satisfying \eqref{s-def}}, \forall f \in nz^* \}.
$
Therefore,  $1-P_1\leq P(A^c\cup V^c)\leq P(A^c)+P(V^c)$.

From the definition of $\kappa$, $\|\bsbSig_{{f}, {nz^*}} \bsbSig_{{nz^*}}^{-1} \bsbs(\hat\bsbb_{nz^*})\|_2  \leq \kappa \sqrt{p_{nz^*}}
 \|\bsbSig_{nz^*}^{-1}\|_{2} N \sqrt{p_{nz^*\cdot 1}} = \kappa p_{nz^*}   / \mu$,
 $\forall f \in z^*$. It follows that
\begin{align}
P(A^c)\leq p_{z^*}P( \|\bsbe_f' \|_2 \geq (1-\kappa p_{nz^*} / \mu)\lambda \tau_0^2/(\sigma\sqrt N)  )\label{boundA}
\end{align}
where $\bsbe' = \bsbX_{z^*}^{'T} \bsbe/(\sigma\sqrt N) \sim N(0, \bsbS_{z^*}/N)$  because  $\bsbX_{z^*}^{'T}\bsbX_{z^*}^{'}=\bsbS_{z^*}$. Define $M:=(1-\kappa p_{nz^*} / \mu)\lambda \tau_0^2/(\sigma \sqrt N)$.
Based on Lemma \ref{B.2} and Lemma \ref{B.3} and the fact that the diagonal entries of $\bsbS_{z^*}=\bsbSig_{{z^*}} - \bsbSig_{{z^*}, {nz^*}} \bsbSig_{{nz^*}}^{-1} \bsbSig_{{nz^*}, {z^*}}$ are all less than or equal to  $N$, we obtain a bound for \eqref{boundA}:
\begin{align}
P(A^c)\leq \frac{e}{4}p_{z^*}M^2e^{-M^2/4}.\label{boundA2}
\end{align}

Next we bound $P(V^c)$. Suppose the spectral decomposition of $\bsbSig_{nz^*}$ is given by
$\bsb{U} \bsb{D} \bsb{U}^T$ with  the $i$th row of $\bsb{U}$ given by $\bsbu_i^T$, then we can
represent $\bsbSig_{nz^*}^{-1}$ as $\left[\bsb{u}_i^T \bsb{D}^{-1} \bsb{u}_j \right]$, and thus $\mbox{diag}(\bsbSig_{nz^*}^{-1})\leq 1/(N \mu)$.  Moreover, from $\| \bsbSig_{nz^*}^{-1}  \bsbs \|_2 \leq \sqrt{p_{nz^*}}/(N\mu)$, $\|[\bsbSig_{nz^*}^{-1}  \bsbs]_f \|_2 \leq \sqrt{p_{nz^*}}/(N\mu)$, $\forall f \in nz^*$. Introduce $\bsbe{''}=\sqrt{\mu N}\bsbSig_{{nz^*}}^{-1} \bsbX_{{nz^*}}^{T} \bsbe/(\sigma )\sim N(0,   \mu N \bsbSig_{{nz^*}}^{-1})$.
From the last two lemmas,
\begin{align*}
 P(V^c)
&\leq  p_{nz^*} P(\|\bsbe_f''\|_2 \geq (\min_{f\in nz^*}\|\bsbb_{f}^*\|_2 - \frac{\lambda \tau_0^2 \sqrt{p_{nz^*}} }{\mu N})\frac{\sqrt{\mu N}}{\sigma}) \\
&\leq  \frac{e}{4}p_{nz^*}L^2e^{-L^2/4},
\end{align*}
where $L:=(\min_{f\in nz^*}\|\bsbb_{f}^*\|_2 - \lambda \tau_0^2 \sqrt{p_{nz^*}}/(\mu N))\frac{\sqrt{\mu N}}{\sigma}$.
\\


The proof of  \eqref{selbndL02} for the hard-ridge thresholding follows similar lines. First, define $\bsbs(\bsbb; \lambda, \eta)$ as
 \begin{align}
\|\bsbs_{{f}}\|_2 \leq  1, \forall f \in z(\bsbb)\ \mbox{ and } \ \bsbs_{{f}} = \frac{\eta}{\lambda}  {\bsbb_{f}}, \forall f\in  nz(\bsbb).  \label{s-def2}
\end{align}
Then similar to  \eqref{thetaeq-2} we have
\begin{align}
\bsbSig \hat\bsbb = \bsbX^T \bsby - \lambda \tau_0^2  \bsbs(\hat\bsbb; \lambda, \eta),  \label{thetaeq-3}
\end{align}
Let $\bsbX_{z^*}^{''T}:={\bsbX_{z^*}^T-\bsbSig_{z^*,nz^*}\bsbSig_{nz^*}^{-1}[\bsbI-\eta(\bsbSig_{nz^*}+\eta\bsbI)^{-1}]\bsbX_{nz^*}^T}$.
To bound $P_{02}$, from Lemma \ref{dawl:lemma1} and \eqref{s-def2}, we write \eqref{thetaeq-3} as
\begin{eqnarray*}
\begin{cases}
\lambda \tau_0^2\bsbs(\hat \bsbb_{z^*}; {\lambda}, {\eta}) + \bsbS_{z^*} \hat \bsbb_{z^*}= \bsbX_{z^*}^{''T}\bsbe+\eta\tau_0^2\bsbSig_{z^*,nz^*}(\bsbSig_{nz^*}+\eta\tau_0^2\bsbI)^{-1}\bsbb_{nz^*}^*,
\\
\hat\bsbb_{nz^*}= (\bsbSig_{nz^*}+\eta \tau_0^2\bsbI)^{-1}\bsbSig_{nz}\bsbb_{nz^*}^* + (\bsbSig_{nz^*}+\eta \tau_0^2\bsbI)^{-1}\bsbX_{nz^*}^T\bsbe,
\end{cases}
\end{eqnarray*}
where we used $\bsbSig_{nz^*}^{-1}(\bsbSig_{nz^*}+\eta \tau_0^2 \bsbI)^{-1}\bsbSig_{nz^*}=(\bsbSig_{nz^*}+\eta\tau_0^2\bsbI)^{-1}$.
It follows that
\begin{align*}
P_{02} \geq   P(\exists \bsbs \mbox{ satisfying  \eqref{s-def2}  s.t. } \hat \bsbb_{z^*} = \bsb{0} \mbox{ and } \|\hat \bsbb_{f}\|_2 \geq \lambda/(1 + \eta), \forall f \in nz^*) \geq P(A \cap V)
\end{align*}
with
\begin{align*}
A & :=  \{ \|\bsbX_{f}^{''T}\bsbe+\eta \tau_0^2\bsbSig_{f,nz^*}(\bsbSig_{nz^*}+\eta\tau_0^2\bsbI)^{-1}\bsbb_{nz^*}^*\|_2\leq \lambda \tau_0^2, \forall f\in z^*\}, \\
V &:= \{\|[(\bsbSig_{nz^*}+\eta\tau_0^2\bsbI)^{-1}\bsbSig_{nz^*}\bsbb_{nz^*}^*]_f +[(\bsbSig_{nz^*}+\eta\tau_0^2\bsbI)^{-1}\bsbX_{nz^*}^T\bsbe]_f\|_2\geq \frac{\lambda}{1+\eta}, \forall f\in nz^*\}.
\end{align*}
For  $\bsbe'=\bsbX_{z^*}^{''T}\bsbe/(\sigma \sqrt N)$ and $\bsbe''=\frac{\mu N+\eta\tau_0^2}{\sqrt{\mu N}}(\bsbSig_{nz^*}+\eta\tau_0^2\bsbI)^{-1}\bsbX_{nz^*}^T\bsbe/\sigma$, their covariance matrices are computed as
$$(\bsbSig_{z^*} - \bsbSig_{z^*,nz^*}[\bsbI-\eta^2(\bsbSig_{nz^*}+\eta\bsbI)^{-2}]\bsbSig_{nz^*}^{-1}\bsbSig_{z^*,nz^*}^T)/N$$
and $$\frac{(\mu N+\eta \tau_0^2)^2}{\mu N}(\bsbSig_{nz^*}+\eta \tau_0^2\bsbI)^{-1} \bsbSig_{nz^*} (\bsbSig_{nz^*}+\eta \tau_0^2\bsbI)^{-1},$$ respectively. Furthermore, it is not difficult to see that  their diagonal entries are  bounded by 1 (under $\eta \leq \mu N/\tau_0^2$). Therefore,
\begin{align*}
1-P_{02} &  \leq   P(\exists f\in z^* s.t. \|\bsbe_f^{'} \|_2 \geq  \frac{1}{\sigma \sqrt N}(\lambda \tau_0^2- \frac{\eta \tau_0^2 \kappa \sqrt{p_{nz^*}} \|\bsbb_{nz^*}^*\|_2}{\mu N+\eta \tau_0^2} )  )  \\
& \qquad+ P(\exists f\in nz^* s.t. \|\bsbe_f^{''}\|_2 \geq (\iota - \frac{\lambda}{1 + \eta})\frac{\mu N+ \eta \tau_0^2}{\sqrt{\mu N}\sigma} )
 \\
& \leq\frac{e}{4} p_{z^*} M'^2 e^{-M'^2/4}+\frac{e}{4} p_{z^*} L'^2 e^{-L'^2/4},
\end{align*}
where $M':= \frac{1}{\sigma \sqrt N}(\lambda \tau_0^2 - \frac{\eta \tau_0^2}{\mu N+\eta \tau_0^2} \kappa \sqrt{p_{nz^*}} \|\bsbb_{nz^*}^*\|_2)$ and $L':=(\iota - \frac{\lambda}{1 + \eta})\frac{\mu N+ \eta \tau_0^2}{\sqrt{\mu N}\sigma}$.



\section{Proof of Theorem \ref{th:scr}}
\label{app:proofscrconv}

We show the proof for the group form only. The proof in the non-group case is similar (and simpler).
First, we introduce a group {quantile thresholding rule} $\vec\Theta^{\#}(\cdot; {m}, \eta)$ as a variant of the hard-ridge thresholding. Given $1\le {m}  \le |\mathcal F|$ and $\eta\ge 0$, $\Theta^{\#}(\cdot;{m}, \eta): \bsb{a}\in {\mathbb R}^{2D}\rightarrow \bsb{b}\in {\mathbb R}^{2D}$ is defined as follows:   $\bsb{b}_f = \bsb{a}_f/(1+\eta)$ if $\|\bsb{a}_f\|_2$ is among the $m$ largest norms in the set of $\{\|\bsb{a}_f\|_2: f\in \mathcal F\}$, and $\bsb{b}_f=\bsb{0}$ otherwise.
In the case of ties, a random tie breaking rule is used. With the notation, $\bsbb^{(j+1)} = \vec\Theta^{\#}(\bsbxi^{(j+1)}; m, \eta)$.

From the algorithm, $nz(\bsbb^{(j)})\leq m$ is obvious. To prove the function value decreasing property, we introduce the following lemma.
\begin{lemma}
$\hat \bsbb  = \vec\Theta^{\#}(\bsbxi; m, \eta)$ is a globally optimal solution to
\begin{align}
\min_{\bsbb} \frac{1}{2} \| \bsbxi - \bsbb\|_2^2  + \frac{\eta}{2} \|\bsbb\|_2^2=:f_0(\bsbb;\eta) \quad \mbox{ s.t. } nz(\bsbb) \leq m. \label{opt_simp}
\end{align}\label{lm:1}
\end{lemma}
Let $I\subset \mathcal F$ with $|I|=m$. Assuming  $\bsbb_{I^c}=\bsb{0}$, we get the optimal solution $\hat\bsbb$ with     $\hat\bsbb_{I}=\bsbxi_I/(1+\eta)$. It follows that  $f_0(\hat \bsbb; \eta)= \frac{1}{2} \|\bsbxi\|_2^2 - \frac{1}{2(1+\eta)} \sum_{f \in I} \|\bsbxi_f\|_2^2$. Hence the group quantile thresholding $\vec\Theta^{\#}(\bsbxi; m, \eta)$ yields a global minimizer.

Based on this lemma, \eqref{fundecrease} can be proved following the lines of Appendix \ref{app:proofconv}. Details are omitted. 
\\

Now suppose that $\eta>0$ and  the  following \textbf{no tie occurring assumption} holds:\\
\emph{Assumption $\mathfrak{B}$}:   \emph{No ties occur in performing $\vec\Theta^{\#}(\bsbxi; m, \eta)$ for any $\bsbxi$ in the closure of $\{\bsbxi^{(j)}\}$, i.e., either $\|\bsbxi_{(m)}\|_2 > \|\bsbxi_{(m+1)}\|_2 $ or $\|\bsbxi_{(m)}\|_2 = \|\bsbxi_{(m+1)}\|_2=0 $ occurs, where $\|\bsbxi_{(m)}\|_2$ and $\|\bsbxi_{(m+1)}\|_2$ are the $m$-th and $(m+1)$-th largest norms in $\{\|\bsbxi_f\|_2: f\in \mathcal F\}$. }

From  $P(\bsbb^{(j+1)}; \eta) \leq G(\bsbb^{(j+1)}, \bsbxi^{(j+1)}, \bsbb^{(j+1)}, \bsbxi^{(j+1)}; \eta)\leq G(\bsbb^{(1)}, \bsbxi^{(1)}, \bsbb^{(1)}, \bsbxi^{(1)}; \eta)$ and $\eta>0$, $\bsbb^{(j)}$ is uniformly bounded. Let $\bsbb^{\circ}$ be a limit point of $\bsbb^{(j)}$ satisfying  $\bsbb^{\circ}=\lim_{l\rightarrow \infty} \bsbb^{(j_l)}$. Then from
$$G(\bsbb^{(j_l)}, \bsbxi^{(j_l)}, \bsbb^{(j_l)},  \bsbxi^{(j_l)}) - G(\bsbb^{(j_l+1)}, \bsbxi^{(j_l+1)}, \bsbb^{(j_l+1)},  \bsbxi^{(j_l+1)})
\rightarrow 0,$$
 we have  $\bsbxi^{(j_l)} - \bsbxi^{(j_l+1)} \rightarrow 0$ and thus $\bsbb^{(j_l)} - \bsbb^{(j_l+1)}\rightarrow 0$.
That is,
$(1-\omega)\bsbxi^{(j_l)}+\omega ( \bsbb^{(j_l)}+ \bsbX^T(\bsby-\bsbX\bsbb^{(j_l)}) ) - \bsbxi^{(j_l)}  \rightarrow 0$ and
$\vec\Theta^{\#} (\bsbxi^{(j_l)}; m, \eta)-\bsbb^{(j_l)}\rightarrow 0$. We obtain
$\bsbb^{\circ} = \lim_{l\rightarrow \infty} \vec\Theta^{\#} ( \bsbb^{(j_l)} + \bsbX^T (\bsby - \bsbX\bsbb^{(j_l)}); m, \eta).$
Because the limit of $\bsbb^{(j_l)} + \bsbX^T (\bsby - \bsbX\bsbb^{(j_l)})$ exists, it is easy to show that $\bsbb^{\circ}$ is a fixed point of
$$
\bsbb = \vec\Theta^{\#} ( \bsbb + \bsbX^T (\bsby - \bsbX\bsbb); m, \eta).
$$
Let $nz^{\circ}=nz(\bsbb^{\circ})$.
By the definition of $\vec\Theta^{\#}$,
$\bsbb_{nz^{\circ}}^{\circ} = \bsbb_{nz^{\circ}}^{\circ}/(1+\eta) +  \bsbX_{nz^{\circ}}^T  (\bsby -  \bsbX \bsbb^{\circ} )/(1+\eta),$
and thus  $\eta \bsbb_{nz^{\circ}}^{\circ}+ \bsbX_{nz^{\circ}}^T   (\bsby - \bsbX_{nz^{\circ}} \bsbb_{nz^{\circ}}^{\circ}) =\bsb{0}$. But this is the KKT equation of the convex optimization problem \begin{align}
\min_{\bsbg} \frac{1}{2} \|\bsby - \bsbX_{I}\bsbg\|_2^2 + \frac{\eta}{2} \|\bsbg\|_2^2 \label{localridge}
\end{align}
 with  $I=nz^{\circ}$ given. Therefore,  $\bsbb^{\circ}$ is the   ridge regression estimate restricted to $\bsbX_{nz^{\circ}}$. Note that $\eta>0$ guarantees its uniqueness given $I$.

Next, based on Ostrowski's convergence theorem, the boundedness of $\bsbb^{(j)}$ and $\lim \|\bsbb^{(j)}- \bsbb^{(j+1)}\| = 0$ imply that  the set of limit points of $\bsbb^{(j)}$ (denoted by $L$) must be connected. On the other hand, the set of all restricted ridge regression estimates (denoted by $R$) is finite.
Therefore,  $\lim \bsbb^{(j)}=\bsbb^{\circ}$. The convergence of $\bsbxi^{(j)}$ is guaranteed as well.

\vspace{-.1in}
\small{
\bibliographystyle{IEEEtran}
\bibliography{specanal}

\begin{thebibliography}{10}
\providecommand{\url}[1]{#1}
\csname url@samestyle\endcsname
\providecommand{\newblock}{\relax}
\providecommand{\bibinfo}[2]{#2}
\providecommand{\BIBentrySTDinterwordspacing}{\spaceskip=0pt\relax}
\providecommand{\BIBentryALTinterwordstretchfactor}{4}
\providecommand{\BIBentryALTinterwordspacing}{\spaceskip=\fontdimen2\font plus
\BIBentryALTinterwordstretchfactor\fontdimen3\font minus
  \fontdimen4\font\relax}
\providecommand{\BIBforeignlanguage}[2]{{%
\expandafter\ifx\csname l@#1\endcsname\relax
\typeout{** WARNING: IEEEtran.bst: No hyphenation pattern has been}%
\typeout{** loaded for the language `#1'. Using the pattern for}%
\typeout{** the default language instead.}%
\else
\language=\csname l@#1\endcsname
\fi
#2}}
\providecommand{\BIBdecl}{\relax}
\BIBdecl

\bibitem{stoica2005spectral}
P.~Stoica and R.~Moses, \emph{{Spectral Analysis of Signals}}.\hskip 1em plus
  0.5em minus 0.4em\relax Pearson/Prentice Hall, 2005.

\bibitem{stoica2009spectral}
P.~Stoica, J.~Li, and H.~He, ``{Spectral analysis of nonuniformly sampled data:
  a new approach versus the periodogram},'' \emph{IEEE Transactions on Signal
  Processing}, vol.~57, no.~3, pp. 843--858, 2009.

\bibitem{lomb1976least}
N.~Lomb, ``{Least-squares frequency analysis of unequally spaced data},''
  \emph{Astrophysics and space science}, vol.~39, no.~2, pp. 447--462, 1976.

\bibitem{schmidt1986multiple}
R.~Schmidt, ``{Multiple emitter location and signal parameter estimation},''
  \emph{IEEE Transactions on Antennas and Propagation}, vol.~34, no.~3, pp.
  276--280, 1986.

\bibitem{li2002efficient}
J.~Li and P.~Stoica, ``{Efficient mixed-spectrum estimation with applications
  to target feature extraction},'' \emph{IEEE Transactions on Signal
  Processing}, vol.~44, no.~2, pp. 281--295, 2002.

\bibitem{chen1998application}
S.~Chen and D.~Donoho, ``{Application of basis pursuit in spectrum
  estimation},'' in \emph{Proceedings of the IEEE International Conference on
  Acoustics, Speech and Signal Processing}, vol.~3, 1998, pp. 1865--1868.

\bibitem{li2009mimo}
J.~Li, P.~Stoica, and E.~Corporation, \emph{{MIMO radar signal
  processing}}.\hskip 1em plus 0.5em minus 0.4em\relax Wiley Online Library,
  2009.

\bibitem{bourguignon2007sparsity}
S.~Bourguignon, H.~Carfantan, and J.~Idier, ``{A sparsity-based method for the
  estimation of spectral lines from irregularly sampled data},'' \emph{IEEE
  Journal of Selected Topics in Signal Processing}, vol.~1, no.~4, p. 575,
  2007.

\bibitem{candes2008enhancing}
E.~Candes, M.~Wakin, and S.~Boyd, ``{Enhancing sparsity by reweighted $l_1$
  minimization},'' \emph{Journal of Fourier Analysis and Applications},
  vol.~14, no.~5, pp. 877--905, 2008.

\bibitem{fuchs2002minimal}
J.~Fuchs and B.~Delyon, ``{Minimal $L_1$-norm reconstruction function for
  oversampled signals: applications to time-delay estimation},'' \emph{IEEE
  Transactions on Information Theory}, vol.~46, no.~4, pp. 1666--1673, 2002.

\bibitem{sparspec}
{S. Bourguignon}, {H. Carfantan}, and {T. B\"ohm}, ``Sparspec: a new method for
  fitting multiple sinusoids with irregularly sampled data,'' \emph{Astron.
  Astrophys}, vol. 462, no.~1, pp. 379--387, 2007.

\bibitem{blumensath2009iterative}
T.~Blumensath and M.~Davies, ``{Iterative hard thresholding for compressed
  sensing},'' \emph{Applied and Computational Harmonic Analysis}, vol.~27,
  no.~3, pp. 265--274, 2009.

\bibitem{blumensath2009normalised}
------, ``{Normalized iterative hard thresholding: Guaranteed stability and
  performance},'' \emph{IEEE Journal on Selected Topics in Signal Processing},
  vol.~4, no.~2, pp. 298--309, 2009.

\bibitem{lza-f}
M.~Hyder and K.~Mahata, ``An $l_0$ norm based method for frequency estimation
  from irregularly sampled data,'' in \emph{Proceedings of IEEE ICASSP}, 2010,
  pp. 4022--4025.

\bibitem{isl0}
------, ``An improved smoothed approximation algorithm for sparse
  representation,'' \emph{IEEE Transactions on Signal Processing}, vol.~58,
  no.~4, pp. 2194 --2205, april 2010.

\bibitem{donoho2006stable}
D.~Donoho, M.~Elad, and V.~Temlyakov, ``{Stable recovery of sparse overcomplete
  representations in the presence of noise},'' \emph{IEEE Transactions on
  Information Theory}, vol.~52, no.~1, pp. 6--18, 2006.

\bibitem{candes2006stable}
E.~Cand{\`e}s, J.~Romberg, and T.~Tao, ``{Stable signal recovery from
  incomplete and inaccurate measurements},'' \emph{Communications on Pure and
  Applied Mathematics}, vol.~59, no.~8, pp. 1207--1223, 2006.

\bibitem{zhang2008sparsity}
C.~Zhang and J.~Huang, ``{The sparsity and bias of the Lasso selection in
  high-dimensional linear regression},'' \emph{Annals of Statistics}, vol.~36,
  no.~4, pp. 1567--1594, 2008.

\bibitem{zhao2006model}
P.~Zhao and B.~Yu, ``On model selection consistency of lasso,'' \emph{Journal
  of Machine Learning Research}, vol.~7, no.~2, pp. 2541--2563, 2006.

\bibitem{candplan09}
E.~J. Cand{\`e}s and Y.~Plan, ``Near-ideal model selection by {$\ell_1$}
  minimization,'' \emph{Ann. Statist.}, vol.~37, no.~5A, pp. 2145--2177, 2009.

\bibitem{scargle1982studies}
J.~Scargle, ``{Studies in astronomical time series analysis. II-Statistical
  aspects of spectral analysis of unevenly spaced data},'' \emph{The
  Astrophysical Journal}, vol. 263, pp. 835--853, 1982.

\bibitem{mallat1993matching}
S.~Mallat and Z.~Zhang, ``{Matching pursuits with time-frequency
  dictionaries},'' \emph{IEEE Transactions on Signal Processing}, vol.~41,
  no.~12, pp. 3397--3415, 1993.

\bibitem{natarajan1995sparse}
B.~Natarajan, ``{Sparse approximate solutions to linear systems},'' \emph{SIAM
  Journal on Computing}, vol.~24, no.~2, pp. 227--234, 1995.

\bibitem{holland1992genetic}
J.~Holland, ``{Genetic algorithms},'' \emph{Scientific American}, vol. 267,
  no.~1, pp. 66--72, 1992.

\bibitem{harikumar1996new}
G.~Harikumar and Y.~Bresler, ``{A new algorithm for computing sparse solutions
  to linear inverse problems},'' in \emph{Proceedings of IEEE ICASSP}, 1996,
  pp. 1331--1334.

\bibitem{Yuan}
M.~Yuan and Y.~Lin, ``Model selection and estimation in regression with grouped
  variables,'' \emph{Journal of the Royal Statistical Society: Series B
  (Statistical Methodology)}, vol.~68, no.~1, pp. 49--67, 2005.

\bibitem{Block2009}
Y.~Eldar, P.~Kuppinger, and H.~B\"olcskei, ``Block-sparse signals: Uncertainty
  relations and efficient recovery,'' \emph{IEEE Transactions on Signal
  Processing}, vol.~58, no.~6, pp. 3042--3054, Jun. 2010.

\bibitem{SheGLMTISP}
Y.~She, ``An iterative algorithm for fitting nonconvex penalized generalized
  linear models with grouped predictors,'' \emph{Computational Statistics \&
  Data Analysis}, vol.~56, pp. 2976--2990, 2012.

\bibitem{JamesStein1961}
W.~James and C.~Stein, ``Estimation with quadratic loss,'' in \emph{Proceedings
  of the 4th Berkeley Symposium on Mathematical Statistics and Probability,
  Vol. I}.\hskip 1em plus 0.5em minus 0.4em\relax University of California
  Press, 1961, pp. 361--379.

\bibitem{zou2005regularization}
H.~Zou and T.~Hastie, ``{Regularization and variable selection via the elastic
  net},'' \emph{Journal of the Royal Statistical Society: Series B (Statistical
  Methodology)}, vol.~67, no.~2, pp. 301--320, 2005.

\bibitem{SheIPOD}
Y.~She and A.~B. Owen, ``Outlier detection using nonconvex penalized
  regression,'' \emph{Journal of the American Statistical Association}, vol.
  106, no. 494, pp. 626--639, 2011.

\bibitem{she2009thresholding}
Y.~She, ``{Thresholding-based iterative selection procedures for model
  selection and shrinkage},'' \emph{Electronic Journal of Statistics}, vol.~3,
  pp. 384--415, 2009.

\bibitem{daubechies2004iterative}
I.~Daubechies, M.~Defrise, and C.~De~Mol, ``{An iterative thresholding
  algorithm for linear inverse problems with a sparsity constraint},''
  \emph{Communications on Pure and Applied Mathematics}, vol.~57, no.~11, pp.
  1413--1457, 2004.

\bibitem{maleki:optimally}
A.~Maleki and D.~L. Donoho, ``Optimally tuned iterative reconstruction
  algorithms for compressed sensing,'' \emph{IEEE Journal of Selected Topics in
  Signal Processing}, vol.~4, no.~2, pp. 330--341, 2010.

\bibitem{she2010spareg}
Y.~She, ``{Sparse regression with exact clustering},'' \emph{Electronic Journal
  of Statistics}, vol.~4, pp. 1055--1096, 2010.

\bibitem{Park07}
M.~Y. Park and T.~Hastie, ``L1-regularization path algorithm for generalized
  linear models,'' \emph{Journal of the Royal Statistical Society: Series B
  (Statistical Methodology)}, vol.~69, no.~4, pp. 659--677, 2007.

\bibitem{chenchen}
J.~Chen and Z.~Chen, ``Extended {Bayesian} information criterion for model
  selection with large model space,'' \emph{Biometrika}, vol.~95, pp. 759--771,
  2008.

\bibitem{SLIM}
X.~Tan, W.~Roberts, J.~Li, and P.~Stoica, ``Sparse learning via iterative
  minimization with application to mimo radar imaging,'' \emph{IEEE
  Transactions on Signal Processing}, vol.~59, no.~3, pp. 1088--1101, 2011.

\bibitem{SLIM-cg}
D.~Vu, L.~Xu, M.~Xue, and J.~Li, ``Nonparametric missing sample spectral
  analysis and its applications to interrupted sar,'' \emph{IEEE Journal of
  Selected Topics in Signal Processing}, vol.~6, no.~1, pp. 1--14, 2012.

\bibitem{fan2008sure}
J.~Fan and J.~Lv, ``{Sure independence screening for ultrahigh dimensional
  feature space},'' \emph{Journal of the Royal Statistical Society: Series B
  (Statistical Methodology)}, vol.~70, no.~5, pp. 849--911, 2008.

\bibitem{iaa-apes}
T.~Yardibi, J.~Li, P.~Stoica, M.~Xue, and A.~B. Baggeroer, ``Source
  localization and sensing: A nonparametric iterative adaptive approach based
  on weighted least squares,'' \emph{IEEE Transactions on Aerospace and
  Electronic Systems}, vol.~46, no.~1, pp. 425--443, 2010.

\bibitem{spice}
P.~Stoica, P.~Babu, and J.~Li, ``New method of sparse parameter estimation in
  separable models and its use for spectral analysis of irregularly sampled
  data,'' \emph{IEEE Transactions on Signal Processing}, vol.~59, no.~1, pp.
  35--47, 2011.

\bibitem{And55}
T.~W. Anderson, ``The integral of a symmetric unimodal function over a
  symmetric convex set and some probability inequalities,'' \emph{Proc. Amer.
  Math. Soc.}, vol.~6, pp. 170--176, 1955.

\bibitem{cava02}
L.~Cavalier, G.~K. Golubev, D.~Picard, and A.~B. Tsybakov, ``Oracle
  inequalities for inverse problems,'' \emph{Ann. Statist.}, vol.~30, no.~3,
  pp. 843--874, 2002, dedicated to the memory of Lucien Le Cam.

\end{thebibliography}
}

\end{document}